\pdfoutput=1

\documentclass[11pt]{article}

\usepackage[final]{acl2023}
\usepackage{times}
\usepackage{latexsym}
\usepackage{multirow}
\usepackage{cite} 
\usepackage[T1]{fontenc}

\usepackage[utf8]{inputenc}

\usepackage{microtype}

\usepackage{inconsolata}
\usepackage{adjustbox}
\usepackage{booktabs}
\usepackage{longtable}
\usepackage{tcolorbox}
\usepackage{stackengine}
\usepackage{multirow}
\usepackage{graphicx}
\usepackage{wrapfig}
\usepackage{hyperref}       
\usepackage{url}            
\usepackage{amsfonts}       
\usepackage{nicefrac}       
\usepackage{microtype}      
\usepackage{xcolor,colortbl}         
\usepackage{times}
\usepackage{latexsym}
\usepackage{multicol}
\usepackage{blindtext}
\usepackage{tabu}
\usepackage{amsmath, bm}

\usepackage{subcaption}
\usepackage{caption}
\usepackage[normalem]{ulem}
\usepackage{soul}
\usepackage[shortlabels]{enumitem}
\usepackage{array}
\usepackage{pgffor}
\usepackage{textcomp}
\usepackage{amssymb}
\usepackage{pifont}
\usepackage{booktabs}
\usepackage{tabularx}
\usepackage{color}
\usepackage{xcolor}
\usepackage{mwe}
\usepackage{float}

\definecolor{lightgreen}{rgb}{0.8, 0.95, 0.8}
\definecolor{lightred}{rgb}{0.95, 0.8, 0.8}
\definecolor{naplesyellow}{rgb}{0.98, 0.85, 0.37}
\definecolor{pastelyellow}{rgb}{0.99, 0.99, 0.59}

\title{Investigating the Shortcomings of LLMs in Step-by-Step Legal Reasoning}

\author{Venkatesh Mishra$^{1*}$ \quad Bimsara Pathiraja$^1$\thanks{\ \ Equal Contribution} \quad Mihir Parmar$^1$ \quad Sat Chidananda$^1$  \\ \textbf{Jayanth Srinivasa}$^2$\quad  \textbf{Gaowen Liu}$^2$ \quad \textbf{Ali Payani}$^2$ \quad \textbf{Chitta Baral}$^1$ \\\\ 
$^1$Arizona State University \quad $^2$Cisco Research\\
\small{\texttt{\{vmishr23, bpathir1, chitta\}@asu.edu}}
}
\begin{document}
\maketitle
\begin{abstract}

Reasoning abilities of LLMs have been a key focus in recent years. One challenging reasoning domain with interesting nuances is legal reasoning, which requires careful application of rules, and precedents while balancing deductive and analogical reasoning, and conflicts between rules.  Although there have been a few works on using LLMs for legal reasoning, their focus has been on overall accuracy. In this paper, we dig deeper to do a step-by-step analysis and figure out where they commit errors. We use the college-level Multiple Choice Question-Answering (MCQA) task from the \textit{Civil Procedure} dataset and propose a new error taxonomy derived from initial manual analysis of reasoning chains with respect to several LLMs, including two objective measures: soundness and correctness scores. We then develop an LLM-based automated evaluation framework to identify reasoning errors and evaluate the performance of LLMs. The computation of soundness and correctness on the dataset using the auto-evaluator framework reveals several interesting insights. Furthermore, we show that incorporating the error taxonomy as feedback in popular prompting techniques marginally increases LLM performance. Our work will also serve as an evaluation framework that can be used in detailed error analysis of reasoning chains for logic-intensive complex tasks\footnote{Data and source code are available at \url{https://github.com/VenkyMishra/legal_reasoning}}.

\end{abstract}

\section{Introduction}
\begin{figure}[ht]
    \centering     \includegraphics[width=0.8\linewidth]{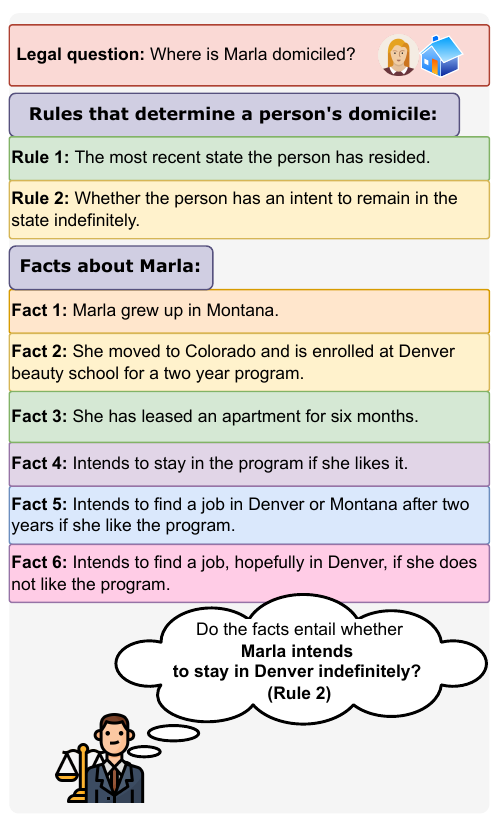}
    \caption{An example of determining domicile in a legal context. A reasoner must discern whether the condition of `indefinite to stay in a place' is met. While many LLMs predict Marla is domiciled in Montana since her program is only for 2 years, legally, her ambiguous plans indicate an intent to remain in Denver indefinitely, making her domiciled in Denver, not Montana.}
    \label{fig:first_teaser}
\end{figure}

\begin{figure*}[t]
    \centering     \includegraphics[width=1.0\linewidth]{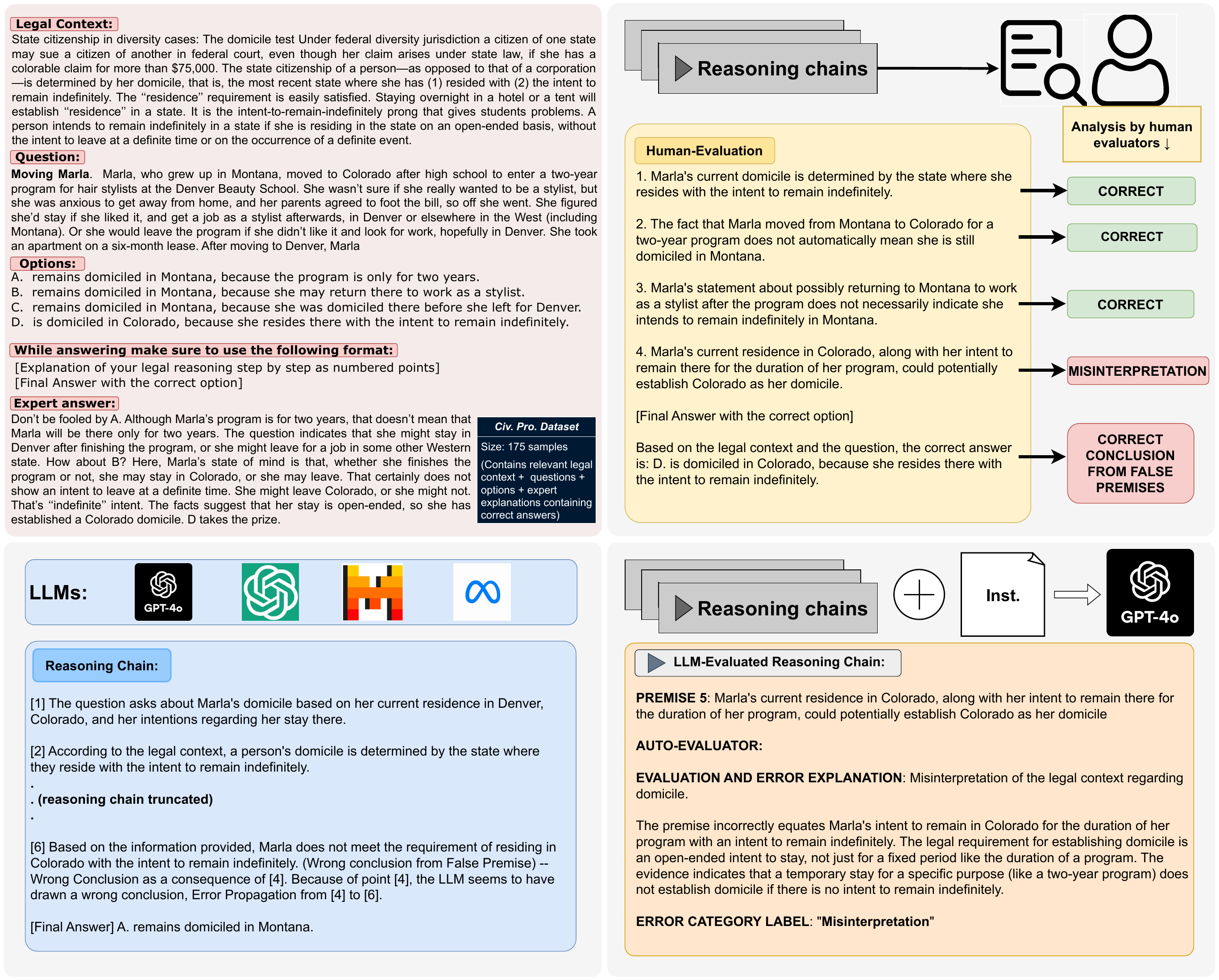}
    \caption{Overview of the proposed pipeline for evaluating legal reasoning in LLMs. The process begins with converting \textit{Civ. Pro.} dataset (\textbf{top left}), followed by generating reasoning chains using LLMs in a zero-shot CoT setting (\textbf{bottom left}). These chains are manually analyzed for various error types (\textbf{top right}), based on the proposed error taxonomy. The pipeline is then automated using an LLM-based system (\textbf{bottom right}) to assess reasoning chains for errors such as misinterpretation, providing insights into the LLMs' reasoning accuracy.}
    \label{fig:teaser}
\end{figure*}

Legal reasoning is a complex process requiring the careful application of rules, and precedents 
while balancing deductive and analogical reasoning, for various legal scenarios \citep{walker2007discovering, hafner2002role}. 
These challenges are heightened by reasoning through uncertainties (Figure \ref{fig:first_teaser}) and ambiguous laws (Figure \ref{fig:legal-reasoning-complexity-part-2}).
In recent years, Large Language Models (LLMs) 
have emerged as the most dominant AI models to process and generate natural language. There has been widespread research to showcase the emergence of natural language understanding (NLU) and reasoning abilities \citep{zelikman2022star, zelikman2024quiet, hao2023reasoning, mondorf2024beyond, lanchantin2024learning} of LLMs. These findings have opened up avenues to utilize LLMs in complex domains like Law \citep{lai2023large}. Consequently, there has been significant research in evaluating the performance of LLMs' legal reasoning ability \citep{guha2023legalbench, blairstanek2023gpt3, kang2023can}. 
{\em In this paper, we focus on providing a detailed, fine-grained analysis of the errors that occur during step-by-step legal reasoning using LLMs.}
While earlier works exist on evaluating step-by-step reasoning of LLMs \citep{golovneva2023roscoesuitemetricsscoring,prasad2023recevalevaluatingreasoningchains}, they do not specifically cater to legal reasoning. 


As shown in Figure \ref{fig:first_teaser}, analyzing legal scenarios requires extensive consideration of critical analysis of prior context. Hence, beyond just evaluating final answers, it is crucial to analyze the step-by-step reasoning chains generated by LLMs and where they falter to gauge their reasoning capabilities to solve such tasks. Despite continued improvements, LLM outputs are affected by fundamental challenges such as hallucinations \citep{dahl2024large, varshney2024investigating}, and misunderstanding long-contexts \citep{lu2024insights}. Our work aims to investigate such errors on a fine-grained level for each step generated to showcase the ability of LLM to perform legal reasoning. To this end, we leveraged the dataset from \citet{bongard2022legalargumentreasoningtask} (referred to as `\textit{Civ. Pro.}' throughout the paper) in the form of MCQA and analyzed reasoning chains generated by LLMs. The dataset comprises of 175 legal scenarios, each providing a comprehensive legal context that includes relevant rules, precedents, and exceptions. Following this context, questions and options are presented to test the understanding and application of these legal principles. The primary goal of using this dataset is to assess how well the LLM reasons logically and contextually when provided with all necessary legal rules, without relying on its inherent knowledge to recall them. This approach highlights the LLM's ability to perform deductive and analogical reasoning in complex, context-rich legal scenarios. Some of the widely used LLMs, Mistral-7B-v2-Instruct, Llama-3-8B-Instruct, GPT-3.5-turbo, GPT-4-turbo, and GPT-4o, are evaluated in zero-shot-CoT setting \citep{wei2023chainofthoughtpromptingelicitsreasoning} (as shown in Figure \ref{fig:teaser}). 

Subsequently, human evaluation of the LLM-generated reasoning chains against the ground truth solutions is performed for a subset of the dataset to find the most common errors being committed in the reasoning steps. This resulted in the development of a detailed error taxonomy, as outlined in Tables \ref{table:premise_errors} and \ref{table:conclusion_errors}, which allowed for a deeper understanding of the underlying causes of LLMs' failures. Additionally, to extend this evaluation across the entire dataset, we develop a method employing `LLMs as auto-evaluators' (inspired by \citet{liu2023gevalnlgevaluationusing, chern2024largelanguagemodelstrusted}) that leverages our proposed error taxonomy. The auto-evaluator framework achieves a recall of $\sim$87.06$\%$ in terms of identifying errors similar to human annotation. To derive better qualitative observations, we utilize two metrics: soundness and correctness (details in \textsection \ref{section: metrics}) to further analyze the reasoning chains. These metrics provide us with interesting insights, the most prominent of them being the inability to generate error-free rationales due to misinterpretation of the contextual nuances.

In the end, we apply various prompting strategies (details in \textsection \ref{section:mitigation_discussion}) known to enhance LLM reasoning abilities with the zero-shot method. We also incorporated our error taxonomy as feedback to the prompting strategies to assess their effectiveness in mitigating errors in legal reasoning. The prompting strategies enhanced with error taxonomy feedback show improved accuracy (maximum of $\sim$4$\%$) which suggests that providing information about errors help in improving the performance of LLMs. We hope that this framework for automatically evaluating step-by-step reasoning in complex tasks will be helpful for future research. In summary, the main contributions of our paper are: 
\begin{enumerate}[noitemsep]
    \item We propose an error taxonomy to systematically identify the most commonly occurring errors in step-by-step legal reasoning.
    \item We develop an LLM-based pipeline to automatically detect errors, and introduce two key metrics—soundness and correctness—to evaluate step-by-step legal reasoning.
    
    \item We investigate the integration of error-taxonomy feedback into various LLM prompting strategies and find that it enhances the LLMs' legal reasoning capabilities.
\end{enumerate}


\label{sec:intro}

\section{Related Work}
\label{sec:related_work}
\paragraph{Legal Reasoning with LLMs}
LegalBench \citep{guha2023legalbench}, LawBench \citep{fei2023lawbench}, LEXGLUE \citep{chalkidis2021lexglue} and LEGALSEMI \citep{kang2024bridging} have introduced exhaustive benchmarks which cover distinct tasks to measure the legal reasoning abilities of LLMs. \citet{dahl2024large} addresses the problem of hallucinations in LLMs, particularly their behavior in generating information that lacks factual accuracy. While these works have evaluated legal reasoning of LLMs on labelled predictions or through expert manual evaluation, our work specifically focuses on evaluating natural language step-by-step rationales with the help of LLMs. We also introduce a task-specific fine-grained error taxonomy to assess the reliability of LLMs in producing error-free reasoning chains.

\paragraph{Evaluation and Verification of Step-by-Step Reasoning Chains through LLMs}
ROSCOE \citep{golovneva2023roscoesuitemetricsscoring} offers a comprehensive suite of metrics to assess various aspects of reasoning quality, including correctness, informativeness, consistency, and coherence. ReCEval \citep{prasad2023recevalevaluatingreasoningchains}, on the other hand, specifically targets incorrect answer detection by specifically analyzing the `correctness' and `informativeness' of reasoning steps. LLM Reasoners \citep{hao2024llmreasonersnewevaluation} introduces a novel evaluation framework for a detailed analysis of large language models’ step-by-step reasoning abilities. \citet{ling2023deductiveverificationchainofthoughtreasoning} introduces Natural Program, a natural language-based deductive reasoning format that decomposes a reasoning verification process into a series of step-by-step subprocesses. \citet{tyen2024llmsreasoningerrorscorrect} explores the limitations of LLMs in detecting reasoning errors and highlights their effectiveness in correcting errors when provided with specific locations. \citet{li2023making} presents the DIVERSE approach involving a three-stage process to identify and correct errors at each step of the reasoning chain. While prior works \citep{patel2024multi, tyagi2024stepbystepreasoningsolvegrid} introduce valuable evaluation frameworks, they are not optimized specifically for a legal reasoning task. Our work utilizes soundness and correctness metrics, which are simple yet effective in offering detailed insights into step-by-step legal reasoning errors. These metrics also make evaluation scalable and reduce the need for manual effort with our formulated LLM-based evaluation framework. 









\begin{table*}[ht]
\small
\centering
\begin{tabular}{p{4cm}|p{11cm}}
\toprule
\textbf{Category}                               & \multicolumn{1}{c}{\textbf{Description}}                                     \\ \midrule
Misinterpretation (associated with Error of Law)                               & The LLM misinterprets or omits some part/entirety of the legal context, question or the options (or a combination of them). This usually leads to the wrong reasoning and selection of wrong conclusion. The following error instances fall under the taxon of misinterpretation: 1. Misunderstanding the legal rules. 2. Misunderstanding the legal situation/issue at hand. 3. Omission of parts of the provided context while reasoning. 4. Incompletely applying a legal rule. 5. Incorrectly applying the legal rule. 6. Wrong assumptions derived from the provided context.                                                                             \\ \midrule
Irrelevant Premise (associated with Error of Law)                             & The LLM generates a premise which is not relevant in solving the question or that it may divert the reasoning chain from solving the question correctly. An premise may be logically valid and factually true but the absence of this premise can still lead to the correct conclusion.                         \\ \midrule
Factual Hallucination (associated with Error of Fact)                          & This error category covers instances where the LLM, during its reasoning process, generates information that is either inconsistent with the facts of the given legal scenario or is entirely fabricated with no basis in reality.                                                                                                                \\ \bottomrule
\end{tabular}
\caption{Error taxonomy for the Premise-level steps. The taxonomy has been developed with consideration for the types of errors that a human reasoner might commit when constructing a rationale for a given legal scenario. Error of Law and Error of Fact are explained in \citep{cornell2024mistake, cornell2024mistake_fact, oreilly2012errors, wilberg2023mistake}. Some fine-grained error instances of the `Misinterpretation' category are shown in Tables \ref{table:human_annotation_example_initial}, \ref{table: misinterpretation-2}, \ref{table: misinterpretation-3} and \ref{table: misinterpretation-4}.}.
\label{table:premise_errors}
\end{table*}
\section{Evaluation of Reasoning Chains}
\label{sec:error_category}

\subsection{The \textit{Civil Procedure} Dataset}
The dataset has been sourced from MCQs present in the `The Glannon Guide To Civil Procedure' \citep{glannon2013guide}. We compile the \textit{Civ. Pro.} dataset with 175 samples of college-level law multiple-choice questions from the US Civil Procedure laws. The questions are primarily designed to evaluate the ability of university-level law students to reason about various legal scenarios about US Civil Procedure laws and provide their final judgment by choosing the most correct option as an answer. The dataset includes relevant legal context, multiple-choice questions, and expert answers with correct explanations provided by legal experts. These elements were extracted and converted into a prompt-based format suitable for LLM inference and the generation of reasoning chains. The \textit{Civ. Pro.} dataset consists of samples comprising of $\mathcal{D} = {<lc_{n}, q_{n}, o_{n}, e_{n}>}$, where $lc_n$, $q_{n}$, $o_{n}$ and $e_{n}$ denote the $n^{th}$ legal context, question, option-set and expert-answer respectively.

 
\subsection{Manual Evaluation Of Reasoning Chains}
\label{section:manual-eval-sec}
Human evaluators are instructed to find flaws in a reasoning chain and explain the flaws in natural descriptive language. To solve a given legal question in \textit{Civ. Pro.}, an LLM generates a set of statements $<$$A: s_1, s_2, ..., s_k, c$$>$, where $A$ represents the legal argument/rationale put forward to solve the problem, with $s_1, s_2...s_k$ being the `\textit{k}' number of intermediate steps generated to reason towards the final conclusion $c$. Each step in the reasoning-chain, including the final conclusion, is separately evaluated for the presence/absence of errors. To create an error taxonomy, we adopt an exhaustive approach, continuously updating the taxonomy until no new errors are identified. Specifically, 120 reasoning chains containing approximately 537 reasoning steps are used for evaluation (generated as responses by four LLMs: Mistral-7B-v2-Instruct, Llama-3-8B-Instruct, GPT-3.5-turbo and GPT-4-turbo, to the same 30 data sample subset). This evaluation helped to solidify our proposed taxonomy as described in \textsection \ref{section:error_taxonomy}. Detailed statistics of the human-evaluations are provided in Tables \ref{table:manual-evaluation-premise} and \ref{table:manual-evaluation-conclusion} of Appendix \ref{section:manual-eval}. Further details regarding annotation guidelines and process, inter-annotator agreement statistics using Cohen's kappa coefficient \citep{cohen1960kappa} and annotation examples are provided in Appendix \ref{section:annotation_guidelines} and Appendix \ref{section:human_annot} (Tables \ref{table:human_annotation_example_initial}-\ref{table:human_annotation_example_final}).



\subsection{Proposed Error Taxonomy}
\label{section:error_taxonomy}
The error taxonomy is designed to mirror the types of errors humans make when reasoning about passage comprehension and constructing rational arguments. It classifies errors into two levels: 1. Premise-level and 2. Conclusion-level errors. Premise-level errors are based on `Errors of Law' and `Errors of Fact' grounded in the legal domain \citep{cornell2024mistake, cornell2024mistake_fact, oreilly2012errors, wilberg2023mistake}. While premise-level errors often influence errors at the conclusion level, many conclusion-level errors occur independently. Conclusion-level errors serve as indicators of the  overall decision-making ability of LLMs in generating the final answer to a legal question. 

\paragraph{Premise-level Errors}
These errors have occurred in one of the premises of the reasoning chain. They highlight the core issue with LLMs that ineffectively prioritize relevant parts of the prior context and incorrectly identify important information. We categorize these errors as shown in Table \ref{table:premise_errors}.
 


\begin{table*}
\small
\begin{tabular}{l|p{3.8cm}|p{8.3cm}}
\toprule
\textbf{Broad-Category}           & \textbf{Sub-Category}                     & \multicolumn{1}{c}{\textbf{Description}}
                                                                                         \\ \midrule
\multirow{4}{*}{Wrong Conclusion} & Wrong Conclusion from False Premise(s)      & This error primarily occurs when the step-by-step rationale generated includes premises that are logically invalid, factually incorrect, irrelevant to solving the question posed, or a combination of these issues.                                                                                 \\ \cmidrule{2-3} 
                                  & Wrong Conclusion from Incomplete Premise(s) & This error occurs when valid and sound premises are provided but fail to fully support the reasoning, leading to the wrong conclusion. A special example is 'Wrong Conclusion from Correct Premises,' where sufficient premises still result in an incorrect conclusion.                 \\ \midrule
Right Conclusion                  & Right Conclusion from False Premise(s)      & This error occurs when the LLM reasons to the correct option while providing a wrong argument. One or multiple premises contain errors which fall under one of the three premise-level error categories and yet lead to the reasoning path choosing the correct option as its final answer. \\ \cmidrule{2-3}
                                  & Right Conclusion from Incomplete Premise(s) & This error occurs when the correct final option is selected, even though the premises provided are incomplete or insufficient to fully justify that conclusion.
                                  
                                                                                                    \\ \cmidrule{2-3}
                                  & Right Conclusion with Hallucinated Content & This error occurs when the LLM selects the correct option but the generated content does not semantically match the provided options. For example, the LLM might output 'Option D. The suspect is \textit{X},' when the actual content is 'Option D. The suspect is \textit{Y} and committed crime \textit{Z},' due to an LLM hallucination. 
                                  \\ \midrule
\end{tabular}
\caption{Error taxonomy for the Conclusion.}
\label{table:conclusion_errors}
\end{table*}





\begin{table}[t]
\small
\scriptsize  
\resizebox{0.9\linewidth}{!}{
\begin{tabular}{p{0.85\columnwidth} }
\toprule
    \textbf{Examples of a reasoning step and final conclusion evaluated by GPT-4o based auto-evaluator} \\ \midrule

       \multirow{3}{*}{}{\textbf{Premise 4:} Marla's Intent: Marla's intent is somewhat ambiguous. She is attending a two-year program and is unsure if she wants to be a stylist. She has considered staying in Denver to work if she likes the program, or leaving the program and looking for work in Denver if she doesn't like it.} \\ \textbf{Auto-Evaluator Evaluation and Error Explanation:} Premise 4: Misinterprets Marla's intent.\\The premise incorrectly suggests ambiguity in Marla's intent. According to the legal context, Marla's plans are open-ended, meaning she does not have definite plans to leave Denver. The ambiguity mentioned does not align with the legal context's explanation of "indefinite intent". \\ \textbf{Error Category Label:} "Misinterpretation" \\ 
    \midrule

        \multirow{4}{*}{}{\textbf{Predicted Conclusion:} Given that Marla resides in Colorado and has no definite plans to leave, she meets the domicile test's requirement of residing with the intent to remain indefinitely.

Final answer: D,}\\ \textbf{Extracted Ground-Truth Conclusion:} D.  is domiciled in Colorado, because she resides there with the intent to remain indefinitely. \\ \textbf{Auto-Evaluator Evaluation and Error Explanation:} Based on the provided conditions and the information given:

\\- The extracted conclusion's final answer is D.
\\- The ground truth option is also D.
\\- The content of the extracted conclusion and the ground truth option match: both state that Marla is domiciled in Colorado because she resides there with the intent to remain indefinitely.
\\- There is an error in the premises (misinterpretation of Marla's intent).\\\\
Given these points, the appropriate classification is:
\\
\textbf{CORRECT CONCLUSION FROM FALSE PREMISES} \\ \textbf{Error Category Label:} "Correct Conclusion from False Premises"\\

\bottomrule
\end{tabular}}
\caption{Example of reasoning step (premise) and conclusion evaluated by LLM-based `Auto-evaluator' (GPT-4o). The error category labels are extracted from the detailed explanations using an LLM prompted to extract error keywords.}
\label{table:example_annot}
\end{table}

\paragraph{Conclusion-level Errors} 
Conclusion-level errors indicate issues with deductive reasoning, reflecting the LLM's ability to follow premises to reach the correct conclusion. They also reveal how much the decision-making process is influenced by intermediate premises in choosing the final answer. We categorize these errors as shown in Table \ref{table:conclusion_errors}.  



\begin{figure*}[t]
    \centering     
    \includegraphics[width=0.95\linewidth]{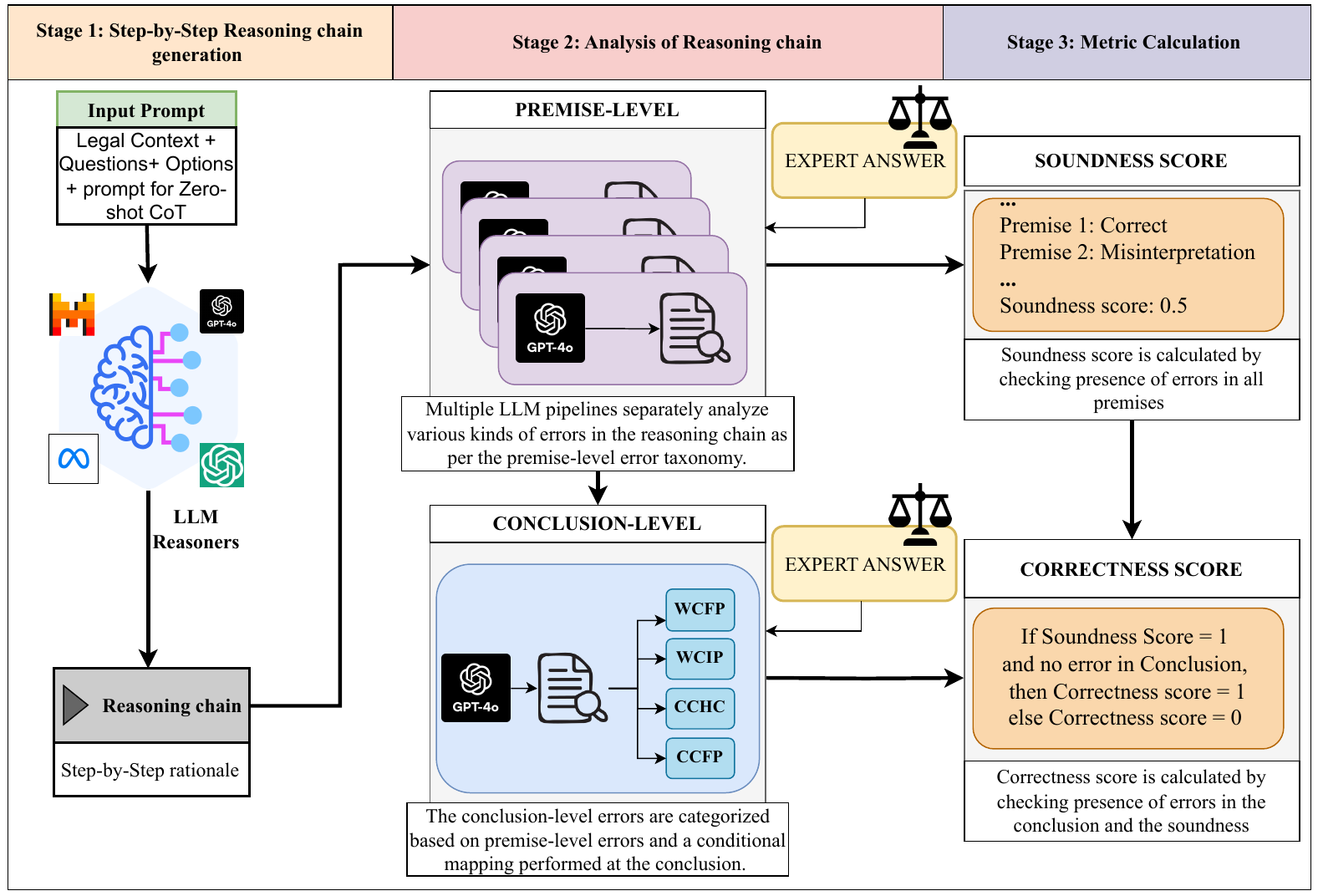}
    \caption{The overall schematic representation of the LLM-based error-detection and evaluation system and the calculation of the metrics. The reasoning chains are produced by 5 LLMs and the expert answer is referenced from the \textit{Civ. Pro.} dataset}
    \label{fig:autoeval}
\end{figure*}

\paragraph{Conclusion from Incomplete Premises \textit{vs.} Correct Premises}
We argue that a `Wrong Conclusion from Correct Premises' is essentially a `Wrong Conclusion from Incomplete Premises' because either the premises, though correct, are incomplete and lead to a wrong conclusion, or the LLM fails to explicitly generate a key premise. This poses a challenge for LLM-based auto-evaluators, as discussed in \textsection \ref{section:llm_aided_eval}, which struggle to assess whether the rationale is sufficient or inadequate.

\subsection{LLM-aided Automatic Evaluation}
\label{section:llm_aided_eval}
Manual analysis of reasoning chains provided a detailed categorization of errors; however, it was time-consuming and, therefore, challenging to scale for the entire dataset. Thus, we develop an alternate approach to leverage LLMs to evaluate the errors in the reasoning chains akin to human evaluation. Specifically, we use GPT-4o as the LLM backbone of the `auto-evaluator' system to identify and label the errors. The auto-evaluator assesses a total of 875 reasoning chains, encompassing approximately 4,844 individual reasoning steps, which include both premise-level and conclusion-level steps (refer Table \ref{table:average_steps}). The details of the implementation are described in Appendix \ref{section:auto-eval} and an example snippet of LLM-aided annotation is provided in Table \ref{table:example_annot}. We develop two approaches for error evaluation:     

\paragraph{Exact Error Label Match} In this approach, we task the `auto-evaluator' with identifying the exact error category labels which the human evaluators had labeled a particular premise/conclusion of a reasoning chain. Experiments revealed significant mislabeling between the auto-evaluator and human evaluators, with many `Misinterpretation' errors at the premise level being labeled as `Irrelevant Premises' or `Factual Hallucination' (Refer Appendix \ref{section:error_disambiguation}) by the auto-evaluator, and vice versa. Hence, we make changes to the auto-evaluators to include error explanations along with the labels.

Another significant challenge was the low error detection rate of factual hallucinations with the help of single-call LLM auto-evaluators. Motivated by \citet{varshney2023stitchtimesavesnine, dhuliawala2023chainofverificationreduceshallucinationlarge}, we develop a multi-call LLM system, consisting of two separate LLM calls,  in which one LLM call creates verification questions to probe various aspects of a premise and another LLM call answers them citing the provided legal context for factuality. A premise is considered to contain factual hallucination if the answers to any of the verification questions contradicts the content of the premise directly.  

\paragraph{Semantic Error Explanation Match} As an alternative approach to the above problems, we develop a multi-analyzer system consisting of three `single-call' and one `multi-call' LLM-based pipeline focused on providing explanation of errors at the premise-level. A `summarizer' LLM (Refer Appendix \ref{section: summ_agg_llm}) combines the individual analyses of all analyzers into a single error explanation for a premise. This enables the pipeline to detect and label multiple errors in a single premise (e.g., a premise containing both misinterpretation and factual hallucination).

To validate the effectiveness of the auto-evaluator, we sample 120 reasoning chains from the manually evaluated set of four LLMs (Mistral-7B-v2-Instruct, Llama-3-8B-Instruct, GPT-3.5-turbo and GPT-4-turbo). The human evaluators then compare their error category assignments as well as explanations to those provided by the auto-evaluator. The recall percentage of detecting an error at the premise level across four LLMs ranged from 83.87$\%$ to 90.6\%. The recall percentage range for detecting an error-free premise step ranged from 86.17$\%$ to 93.85$\%$. The details of the autoevaluator performance statistics are present in Tables \ref{table:auto-eval-agreement-premise} and \ref{table:auto-eval-agreement-conclusion} of Appendix \ref{section:auto-eval-agree}. Figure \ref{fig:autoeval-system} shows the pipeline of the error detection implemented using GPT-4o. 


\section{Experimental Setup}
\label{sec:experiments}

\subsection{Models}
We evaluate a range of closed-source LLMs including GPT-4-Turbo, GPT-4o, GPT-3.5-turbo, and open-weight models Llama-3-8B-Instruct, and Mistral-7B-v2-Instruct on the `\textit{Civ. Pro.}' dataset using the zero-shot CoT prompting setting
\citep{wei2023chainofthoughtpromptingelicitsreasoning, kojima2023largelanguagemodelszeroshot}. While we also conduct the few-shot CoT prompting by providing 1-3 exemplars for solving a legal scenario to LLMs, we find that the final accuracy (\textsection \ref{section: metrics}) is lower across LLMs when compared to the zero-shot CoT setting (refer Appendix \ref{section:few-shot}). NVIDIA A100 GPUs were used to conduct the inference of open-weight models with a batch size of 1. OpenAI API and Gemini API were used for obtaining inference results from the closed-source models. An example prompt used for these experiments is provided in Figure \ref{fig:legal-reasoning-complexity-part-1}.

\subsection{Metrics}
\label{section: metrics}
\paragraph{Accuracy} We use accuracy to demonstrate the capability of LLMs in solving legal scenarios based on their ability to predict the final answer (the conclusion). To calculate this metric, we use the LLM-generated final options chosen as answers and compare them with the available expert answer. 

\paragraph{Soundness Score} We create a step-wise soundness score metric to check the number of premises which is error-free in terms of the absence of errors as delineated by our established error taxonomy. We computed this score by taking its average across a single reasoning chain. The Soundness score (between 0 and 1) is calculated by:

\[
S = \frac{\text{Number of sound premises}}{\text{Total Number of premises}}
\]

\paragraph{Correctness Score} This metric is calculated to evaluate the condition where the reasoning chain must be both error-free at both premise and conclusion levels. Likewise in the soundness metric, the expert answers to extract the ground-truth options as conclusions. A reasoning chain must be both `sound' (with the Soundness score being equal to 1) and arrive at the correct final option as its conclusion. The Correctness score is calculated as:
\[
C =
\begin{cases} 
1, & \text{if (premises +  conclusion) correct} \\
0, & \text{for all other cases} 
\end{cases}
\]

The results of these metrics calculated for 120 reasoning chains annotated by humans are provided in Table \ref{table:metrics_humans} (Appendix \ref{section: metrics_results_humans}). Table \ref{table:metrics} and Figure \ref{fig:acc_correctness_results} show the metric results calculated from auto-evaluator annotations on the entire \textit{Civ. Pro.} dataset.

\section{Results and Analysis}
\label{sec:results}

\subsection{Objective Evaluation}

\paragraph{Soundness metrics are high but correctness scores are low}
Table \ref{table:metrics} shows that the majority of the premises are error-free (with the highest being GPT-4o having 78.4\% of the generated premises being error-free). In contrast, Figure \ref{fig:prem_conc_corr} reveals that an average of $\sim$96\% of reasoning chains leading to conclusions from false premises have one or more misinterpretation errors in the intermediate premises. This finding, aided by empirical human analysis, suggests that much of the LLM-generated reasoning chain re-iterates existing context, while most errors occur in the smaller portion where new `decision-making' inferences are generated. The similarity in correctness score in Mistral-7B-v2-Instruct and Llama-3-8B-Instruct in contrast to the higher accuracy of Llama-3-8B-Instruct could be attributed to the lesser number of steps (see Table \ref{table:average_steps} (Appendix \ref{section:avg_steps_generated})) on average in the reasoning chain of Llama-3-8B-Instruct when compared to Mistral-7B-v2-Instruct. 
\begin{table}[!htbp]
\centering
\begin{tabular}{lccc}
\toprule
Model   & S ($\uparrow$) & A ($\uparrow$)  & C ($\uparrow$)\\ 
\midrule
Mistral-7B-v2-Instruct   & 0.623  & 0.371 & 0.131\\ 
Llama-3-8B-Instruct   & 0.493  & 0.451 & 0.137\\ 
GPT-3.5-turbo   & 0.607  & 0.417 & 0.217\\ 
GPT-4-turbo     & 0.738  & 0.725 & 0.417\\ 
GPT-4o   & \textbf{0.784}  & \textbf{0.737} & \textbf{0.445}\\ 
\bottomrule
\end{tabular}
\caption{The results for soundness, accuracy, and correctness metrics for all LLMs on the \textit{Civ. Pro.} dataset. Here `S' denotes the Soundness, `A' denotes the Accuracy, and `C' denotes the Correctness.}
\label{table:metrics}
\end{table}

\begin{figure}[h]
    \centering
    \includegraphics[width=1.0\linewidth]{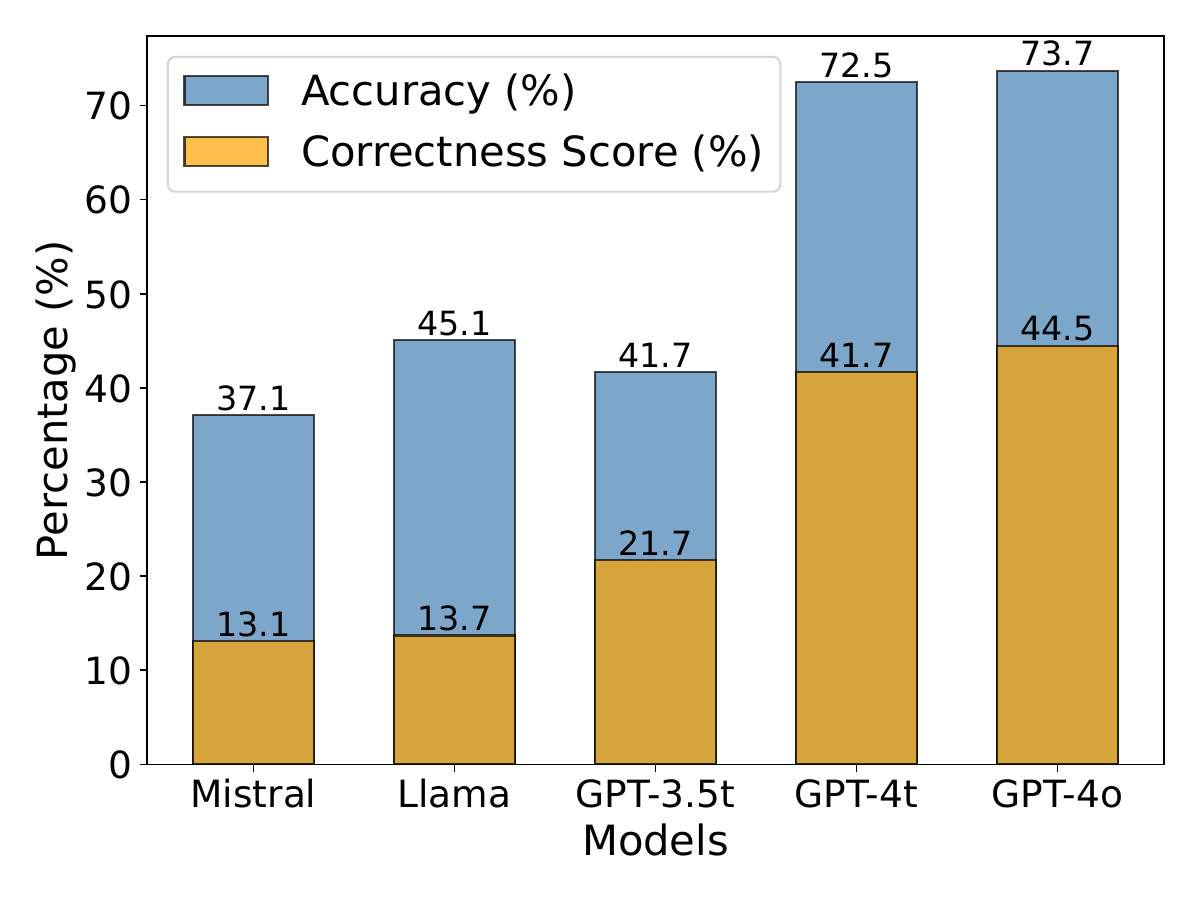}
    \caption{Performance of 5 LLMs in terms of Accuracy vs. Correctness on the \textit{Civ. Pro.} dataset. Here, Mistral stands for Mistral-7B-v2-Instruct, Llama stands for Llama-3-8B-Instruct, GPT-3.5t and GPT-4t stand for GPT-3.5-turbo and GPT-4-turbo respectively.}
    \label{fig:acc_correctness_results}
\end{figure}

\paragraph{Accuracy vs. Correctness Score} Table \ref{table:metrics} and Figure \ref{fig:acc_correctness_results} show a sharp decrease (an average of $\sim$27\%) in the scores of accuracy to correctness across all LLMs. The highest fall in percentage is observed in Llama-3-8B-Instuct (31.4\% decrease). This is significant as it shows that while LLMs can arrive at the correct conclusion, there are a lot of cases where the reasoning chain they generate is not entirely error-free. These results also suggest that LLMs often rely on superficial correlations and patterns, likely learned in the training stages, to arrive at correct conclusions, rather than through genuine reasoning. In high-stakes domains such as legal, financial, and medical fields, it is imperative that the reasoning generated by LLMs is completely error-free as even minor inaccuracies in these critical areas can lead to significant consequences. This also underscores the necessity for robust evaluation mechanisms to ensure the reliability and correctness of model outputs.   

\begin{table}[ht]
\small
\centering
\begin{tabular}{lcccc}
\toprule
Prompting               & B & PS & SC & SD \\
\midrule
\textbf{Gemini} & \multicolumn{4}{c}{} \\
w/o feedback     & 63.31    & 59.17          & \textbf{61.54}        & \textbf{64.50}          \\
w/ feedback      & \textbf{64.50}    & \textbf{62.13}          & 59.76        & 63.31          \\
\midrule
\textbf{Llama} & \multicolumn{4}{c}{} \\
w/o feedback     & 53.71    & 50.29          & 48.58        & 47.42          \\
w/ feedback      & \textbf{57.14}    & \textbf{52.00}          & \textbf{52.57  }      & \textbf{49.14}          \\
\bottomrule
\end{tabular}
\caption{Comparison of accuracy metric for models under different prompting strategies with and without feedback. The models are Gemini-1.5-Flash and Llama-3-8B-Instruct. The prompting strategies abbreviations stands for B: Baseline (CoT), PS: Plan-and-Solve, SC: Self-Correct, and SD: Self-Discovery.}
\label{table:prompt_accuracy}
\end{table}
\paragraph{Larger, proprietary models `reason' better than smaller, open-source models}
Figure \ref{fig:acc_correctness_results} and Table \ref{table:metrics} convey that proprietary models generate more error-free reasoning steps and arrive at the correct conclusion more often than the open-source LLMs. An exception is GPT-3.5-turbo, which performs comparably to Llama-3-8B-Instruct and Mistral-7B-v2-Instruct, suggesting that training data and methods might play a more significant role in enhancing reasoning than merely scaling model parameters. 

\subsection{Reasoning Chain Evaluation}
\paragraph{`Misinterpretations' are the dominant category of errors at premise-level}
Figure \ref{fig:prem-level-cat} (Appendix \ref{section: premise-level-error-dist}) and Figure \ref{fig:prem_conc_corr} reveal that `Misinterpretation' is the most dominant category of error which occurs in the reasoning chains at the premise-level. This highlights that  LLMs struggle to fully grasp the nuanced complexities of legal scenarios requiring the demonstration of critical analysis in zero-shot CoT settings.  

\paragraph{`Wrong Conclusion from False Premises' is the dominant category of error at conclusion-level} The prevalence of `Wrong Conclusion from False Premises' (Figure \ref{fig:conc-level-cat} (Appendix \ref{section: conclusion-level-error-dist})) in conclusion-level errors results from premise-level mistakes leading to incorrect conclusions. However, in GPT-4-turbo and GPT-4o, the dominant error is `Correct Conclusion from False Premises,' suggesting these models may be relying on patterns of similar examples from their training.  

\subsection{Discussion on Error-Mitigation Strategies}
\label{section:mitigation_discussion}
We carry out several experiments on the \textit{Civ. Pro.} dataset, employing widely used prompting techniques alongside the most frequently observed errors we found through \textsection \ref{section:error_taxonomy} with the aim to explore the possibility of enhancing the reasoning capabilities of both closed-source and open-source LLMs. Four prompting techniques are utilized: (1) Chain-of-Thought \citep{wei2022chainofthought} (2) Plan-and-Solve \citep{wang2023plan} (3) Self-Correct  \citep{zhang2024smalllanguagemodelsneed} and (4) Self-Discovery  \citep{zhou2024selfdiscoverlargelanguagemodels}. These techniques are tested with and without incorporating error definitions as feedback, following the Feedback-Learning method \citep{tyagi2024stepbystepreasoningsolvegrid}. Detailed descriptions of the prompting strategies can be found in Appendix \ref{section:prompting_strategies}.

The error definitions are provided in three styles: generic, short, and long. The generic version uses the error definitions from the Feedback-Learning method, while the short and long versions are derived from the error taxonomy described in \textsection\ref{section:error_taxonomy}. All experiments are conducted in a zero-shot setting, and we evaluate each prompting technique based on the accuracy metric. We test one closed-source model, Gemini-1.5-Flash, and one open-source model, Llama-3-8B-Instruct.

As shown in Table \ref{table:prompt_accuracy}, adding the error definitions as feedback showed improvement in accuracy up to ~4\%. For Llama-3-8B-instruct, accuracy improved across all prompting techniques, whereas for Gemini-1.5-Flash, the accuracy increased only for the Chain-of-Thought and Plan-and-Solve methods. From our observations, the decrease in accuracy for these strategies with Gemini resulted due to self-doubting \citep{krishna2023intersectionselfcorrectiontrustlanguage} nature of LLMs. These findings suggest that while feedback on errors provides marginal improvements in LLM performance, there is a need to develop more effective frameworks beyond prompting, such as agent-based methods, that account for these errors and enhance the model’s legal reasoning capabilities.

\section{Conclusion}
Through our work, we assess the reasoning capabilities of LLMs by examining their performance on the `\textit{Civ. Pro.}' dataset designed to evaluate legal reasoning. The sequential workflow of manually evaluating LLM-generated reasoning chains on data samples, creating an error taxonomy based on the commonly occurring errors, automating the error evaluation through LLM-based pipelines lead us to some interesting insights. LLMs still struggle with producing error-free rationales while reasoning about legal scenarios despite being provided with necessary legal knowledge and context. Additionally, we present our findings on implementing various prompting techniques augmented with feedback from our error taxonomy. We hope that our work lays a solid foundation for developing a framework to critically evaluate complex reasoning tasks, such as legal reasoning, and is extendable to other domains in a similar manner.    

\section*{Limitations}
While the \textit{Civ. Pro.} dataset provides a valuable benchmark for assessing the legal reasoning capabilities of LLMs, real-world legal reasoning—particularly in the context of legal judgment prediction—is inherently ambiguous. Legal cases often evolve over time, with new information emerging throughout the litigation process. Consequently, evaluations based on static datasets that capture information at a single point in time may not fully reflect the dynamic nature of legal decision-making, limiting their effectiveness in assessing real-world legal reasoning. There are many legal scenarios in this dataset that are complex in terms of being ambiguous and tricking the reader/reasoner and would probably require legal expertise and experience to understand the nuances required to solve such scenarios. Although our study intends to capture the errors committed by natural-language-based reasoners, the systematic evaluation could benefit from converting natural language to formal language through auto-formalization. While our work mainly focuses on the soundness of the reasoning steps of legal reasoning chains, we would like to acknowledge that properties like consistency, coherence, completeness, and clarity are not directly measured in the current work. Furthermore, the current evaluation of reasoning is restricted to English, leaving room to expand this work into a multilingual context.

\section*{Ethics Statement}
 We obtained the necessary permissions to use the dataset provided by \citet{bongard2022legalargumentreasoningtask}. We have utilized AI assistants, specifically Grammarly and ChatGPT, to correct grammatical errors and rephrase sentences.

\section*{Acknowledgement}

We thank the anonymous reviewers for their constructive suggestions. We extend our gratitude to the Research Computing (RC), and Enterprise Technology at ASU for providing computing resources, and access to the ChatGPT enterprise version for experiments. This work was in part supported by a gift award from Cisco Research. We thank Nimeesh Mahajan, Alexander Carr and Ren Cheng for their help in human evaluations.




\bibliography{references}
\bibliographystyle{acl_natbib}

\clearpage

\appendix
\section{Metrics Results from human evaluations}
\label{section: metrics_results_humans}
Table \ref{table:metrics_humans} shows the statistics of the metrics calculated on the reasoning-chains directly by humans. The results show a marked difference in the values of Accuracy and Correctness across all LLMs manually evaluated by human evaluators. The results in Table \ref{table:metrics}, calculated from the LLM-based auto-evaluator annotations, also reflect the same trends in this table.

\begin{table}[!htbp]
\centering
\small
\begin{tabular}{lccc}
\toprule
LLM                       & S ($\uparrow$)& A ($\uparrow$)& C ($\uparrow$)\\
\midrule
Mistral-7B-v2-Instruct & 0.67      & 0.266    & 0.133       \\
Llama-3-8B-Instruct    & 0.718     & 0.433    & 0.266       \\
GPT-3.5-turbo          & 0.69      & 0.33     & 0.233       \\
GPT-4-turbo            & \textbf{0.748}    & \textbf{0.6}      & \textbf{0.5}\\ \bottomrule       
\end{tabular}
\caption{The results for soundness, accuracy, and correctness metrics for the same 30 LLM reasoning-chain generations across 4 LLMs by human annotators. Here `S' denotes the Soundness, `A' denotes the Accuracy, and `C' denotes the Correctness. The values marked in bold show the highest metric values.}
\label{table:metrics_humans}
\end{table}

\section{Manual Evaluation results}
\label{section:manual-eval}
Tables \ref{table:manual-evaluation-premise} and \ref{table:manual-evaluation-conclusion} show the statistics of the errors found by human evaluators in the premise and conclusion levels based on the process described in \textsection \ref{section:manual-eval-sec} and in accordance to the proposed error taxonomy in \textsection \ref{section:error_taxonomy}.

\begin{table}[!htbp]
\small
\centering
\begin{tabular}{lc}
\toprule
LLM   & Average number of steps\\ 
\midrule
Mistral-7B (893)   & 5.1\\  
Llama-3-8B (642)   & 3.66\\
GPT-3.5-turbo (649)   & 3.70\\
GPT-4-turbo (811)     & 4.63\\ 
GPT-4o (974)   & 5.56\\
\bottomrule
\end{tabular}
\caption{The average number of steps(premises) generated by all LLMs in Zero-shot CoT setting. The numbers in brackets indicate the total number of steps generated by each LLM in the generated reasoning chains (excluding the final conclusion step) for the 175 sample \textit{Civ. Pro.} dataset}
\label{table:average_steps}
\end{table}
\begin{table*}[!htbp]
\small
\centering
\begin{tabular}{lc}
\toprule
LLM   & Error Categories and Frequency\\ 
\midrule
Mistral-7B-v2-Instruct (135 reasoning steps)   & NE - 94, M - 31, FH - 6, IP - 4\\  
Llama-3-8B-Instruct (145 reasoning steps)   & NE - 114, M - 29, FH - 1, IP - 2\\
GPT-3.5-turbo (109 reasoning steps)   & 	NE - 77, M - 30, FH - 1, IP - 0\\
GPT-4-turbo (148 reasoning steps)    & NE - 120, M - 25, FH - 0, IP - 2\\ 
\bottomrule
\end{tabular}
\caption{Statistics on various types of errors identified by human evaluators in the premises of 30 reasoning chains generated by each of the four LLMs. The total number of reasoning steps generated by each LLM is indicated in parentheses. `NE' denotes the absence of errors in the reasoning steps as annotated by human evaluators, `M' represents `Misinterpretation' errors, `FH' indicates `Factual Hallucination,' and `IP' signifies `Irrelevant Premises.'}
\label{table:manual-evaluation-premise}
\end{table*}

\begin{table*}[!htbp]
\small
\centering
\begin{tabular}{lc}
\toprule
LLM   & Error Categories and Frequency\\ 
\midrule
Mistral-7B-v2-Instruct (30 conclusion steps)   & NE - 3, CCFP - 2, CCIP - 1, CCHC - 2, WCFP - 18, WCIP - 3\\  

Llama-3-8B-Instruct (30 conclusion steps)   & NE - 9, CCFP - 4, CCIP - 0, CCHC - 1, WCFP - 13, WCIP - 3\\

GPT-3.5-turbo (30 conclusion steps)   & 	NE - 7, CCFP - 2, CCIP - 0, CCHC - 0, WCFP - 17, WCIP - 4\\

GPT-4-turbo (30 conclusion steps)    & NE - 16, CCFP - 1, CCIP - 0, CCHC - 0, WCFP - 11, WCIP - 1
\\ 
\bottomrule
\end{tabular}
\caption{Statistics on various errors identified by human evaluators in the conclusions of 30 reasoning chains generated by each of the four LLMs. `NE' represents the absence of errors in the conclusion as annotated by human evaluators, `CCFP' denotes `Correct Conclusion from False Premise(s),' `CCIP' indicates 'Correct Conclusion from Incomplete Premise(s),' `CCHC' refers to `Correct Conclusion with Hallucinated Content,' `WCFP' signifies `Wrong Conclusion from False Premise(s),' and `WCIP' represents `Wrong Conclusion from Incomplete Premise(s).'}
\label{table:manual-evaluation-conclusion}
\end{table*}

\section{Percentage distribution of Premise-level errors}
\label{section: premise-level-error-dist}
\begin{figure}[H]
    \centering
    \includegraphics[width=1.0\linewidth]{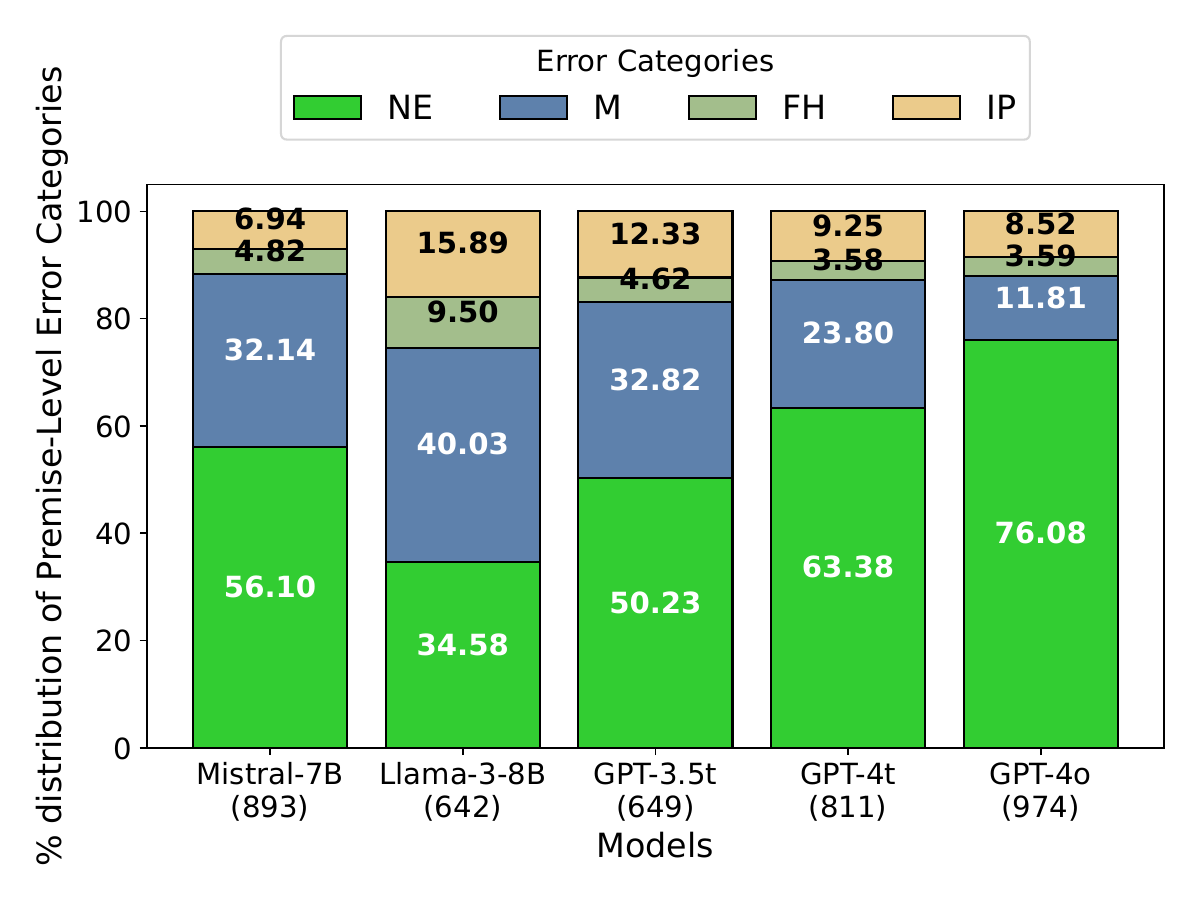}
    \caption{The percentage distribution of the premise-level error categories across the reasoning chains of all 5 LLMs. The total number of steps generated by each model is provided inside the round brackets below the model names. Here ’NE’ denotes Correct Premise (No errors), ’M’ denotes Premise containing a Misinterpretation, ’FH’ denotes Factual Hallucination in the premise, ’IP’ denotes an Irrelevant Premise.}
    \label{fig:prem-level-cat}
\end{figure}

Figure \ref{fig:prem-level-cat} represents the percentage distribution of premise-level errors across the reasoning chains of all
5 LLMs. Due to the lowest number of average steps in reasoning steps, Llama-3-8B-instruct has the highest proportion of errors in the reasoning chains ($\sim$65.4$\%$) containing premise-level errors in the reasoning steps.

\begin{figure}[H]
    \centering
    \includegraphics[width=1.0\linewidth]{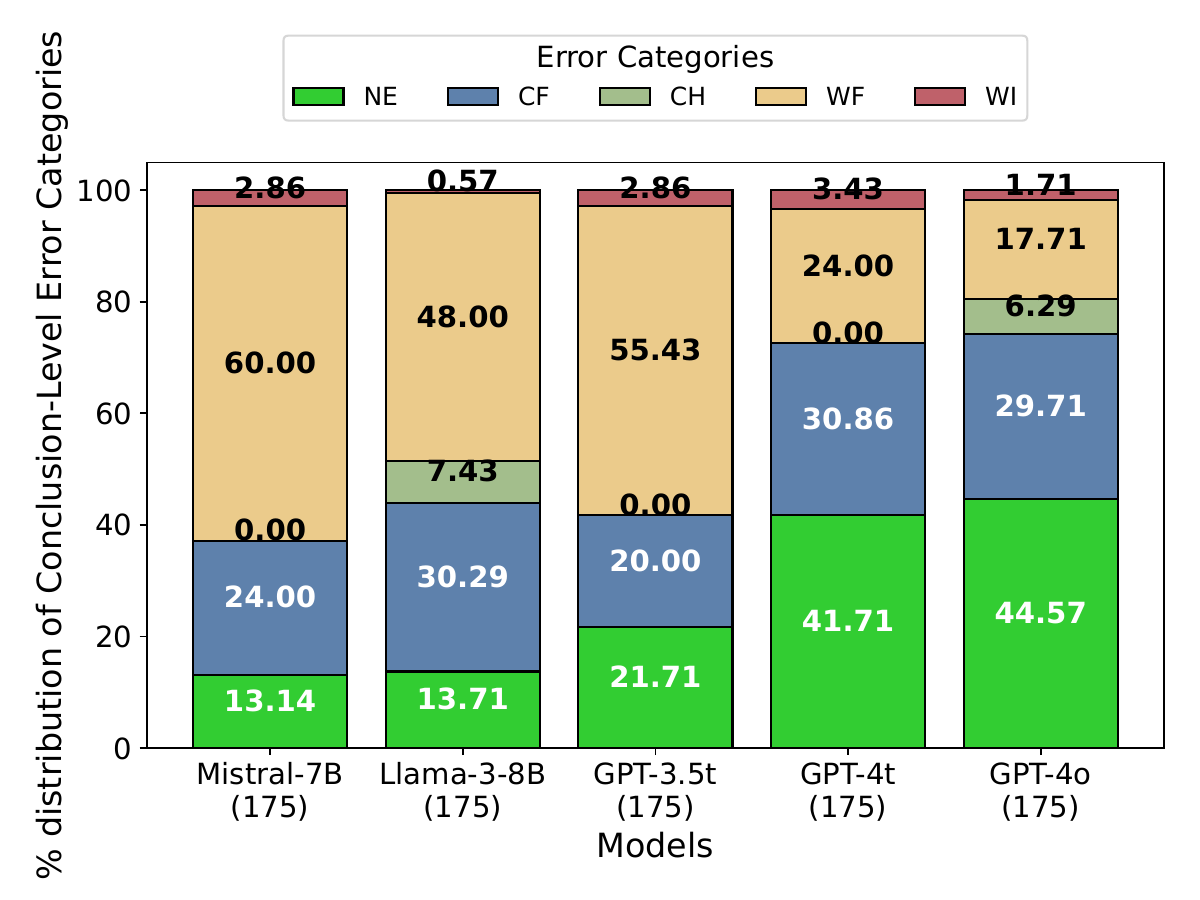}
    \caption{The percentage distribution of the conclusion-level error categories across the reasoning chains of all 5 LLMs. The total number of steps generated by each model is provided inside the round brackets below the model names. Here `NE' denotes Correct Conclusion(CC) From Correct Premises(CP) (No errors), `CF' denotes CC from False Premises (FP), `CH' denotes CC with Hallucinated Content, `WF' denotes Wrong Conclusion(WC) from FP and `WI' denotes EC from Incomplete Premises.}
    \label{fig:conc-level-cat}
\end{figure}

\section{Percentage distribution of Conclusion-level errors}
\label{section: conclusion-level-error-dist}
Figure \ref{fig:conc-level-cat} represents the percentage distribution of conclusion-level errors across the reasoning chains of all
5 LLMs. Mistral-7B-v2-instruct ($\sim$86.8$\%$) and Llama-3-8B-instruct ($\sim$86.3$\%$) have the highest proportion of errors in the reasoning chains containing conclusion-level errors. An interesting observation is that only Llama-3-8B instruct and GPT-4o have non-zero percentages of errors in the category of Correct Conclusion with Hallucinated Content, which could possibly mean that these LLMs have been trained on very similar data to that in the `\textit{Civ. Pro.}' dataset which could be causing these LLMs to spuriously output modified content for the correct options. 

\section{Average number of steps generated by LLMs}
\label{section:avg_steps_generated}
The average number of steps (the premise-level steps) generated by each LLM to solve a legal scenario is provided in Table \ref{table:average_steps}. Llama-3-8B-instruct has the lowest average of number of steps which is a probable cause leading to lower soundness score as shown in Table \ref{table:metrics}. Overall, LLMs mostly try to complete the reasoning chain in around 4-6 steps to arrive at the conclusion. While this is good for not introducing redundancies and keeping the irrelevant premise and hallucination errors low, this could also potentially be an indicator that LLMs do not explicitly output tokens which could be crucial in outlining the reasoning process and making the rationale better in terms of interpretability and explainability.    

\section{Misinterpretations vs. Factual Hallucinations vs. Irrelevant Premises}
\label{section:error_disambiguation}
\begin{figure*}
    \centering     \includegraphics[width=1.0\linewidth]{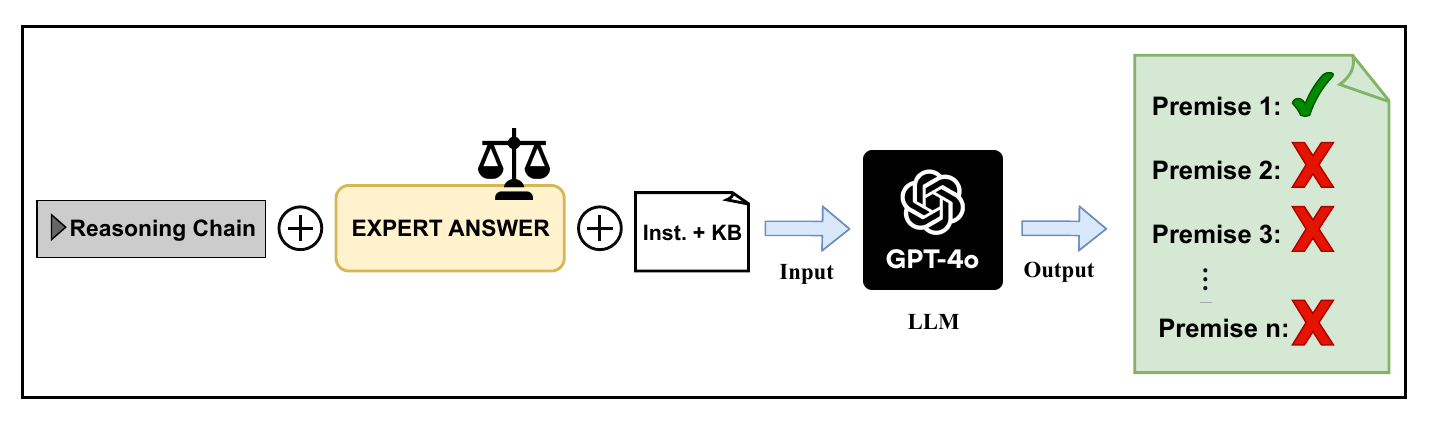}
    \caption{Schematic representation of the `Single-Call' LLM-based auto-evaluator for premise-level error detection. 'Inst + KB' includes instructions for error analysis and the knowledge base with error definitions, based on the proposed error taxonomy. The \textcolor{green}{green} check-mark indicates the absence of errors and the \textcolor{red}{red} cross-mark represents the presence of an error in a premise.}
    \label{fig:autoeval-single-call}
\end{figure*}

A premise is classified as containing a `Misinterpretation' error when the LLM reasoner is making wrong inferences based on the information it generates. There is a `Factual Hallucination' when the information generated by the LLM reasoner (by directly extracting from the provided content of the input legal context, questions, and options) is factually incorrect and can be easily verified while directly going through the input context. A key distinction between Misinterpretation and Factual Hallucination is illustrated in Figures \ref{fig:autoeval-single-call} and \ref{fig:autoeval-multi-call}. Detecting a `Misinterpretation' requires the expert answer to account for complex legal reasoning nuances, while detecting a `Factual Hallucination' does not depend on the expert answer for validation. An irrelevant premise is said to occur when a premise contains unnecessary or tangential information that does not contribute to reasoning toward the correct answer. 

Extensive discussions and iterations occurred to define the taxonomy and differentiate between the three premise-level errors. However, these errors can often overlap or appear together in the same premise, especially in cases where `Misinterpretations' are caused by `Factual Hallucinations'. In the cases of significant overlap, both the human and the auto-evaluator were instructed to annotate multiple errors for the same premise. Detecting multiple errors can be crucial for longer premises which creates inferences on a greater number of contextual factors and nuances and hence are vulnerable to more of errors occurring in them.  

\section{Human Annotation Guidelines and Process}
\label{section:annotation_guidelines}
\begin{figure}[H]
    \centering     \includegraphics[width=0.97\linewidth]{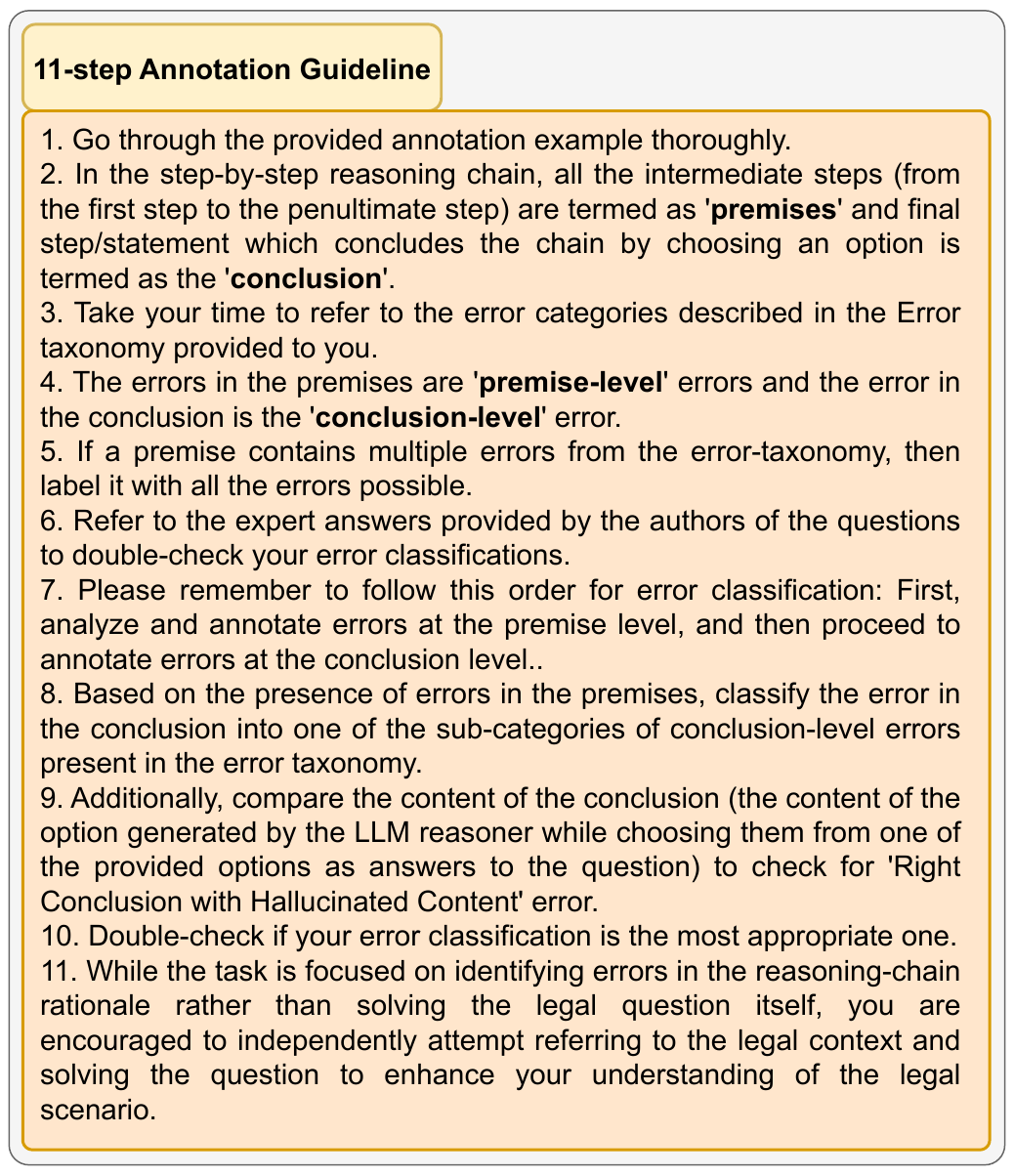}
    \caption{The 11-step guideline provided to the annotators for conducting manual evaluations of the LLM-generated rationale} 
    \label{fig:annotation-guideline}
\end{figure}

A set of 11 annotation guidelines, as detailed in Figure \ref{fig:annotation-guideline}, was developed and provided to annotators for manual evaluation and annotation. Four annotators, two students from undergraduate and graduate level each respectively, participated in this process, and cross-evaluation was conducted at the final stage to resolve any discrepancies. The annotation guidelines in Figure \ref{fig:annotation-guideline} also served as the basis for creating system prompts used in the LLM-based auto-evaluator. 

As an additional study to check for inter-annotator agreement, three annotators separately annotated 10 reasoning chains from Mistral-7B-v2-Instruct and the calculated the Cohen's kappa coefficient ($\kappa$). The $\kappa$ values for the 3 annotator pairs (by selecting 2 unique annotators out of 3 everytime) came out to be 0.862, 0.783 and 0.813, making the average $\kappa$ value to be 0.819. This indicated there is almost perfect agreement between the annotators according to Cohen's kappa metric interpretation. 

\section{Human Annotation Examples}
\label{section:human_annot}
Initially, Mistral-7B-v2-Instruct was selected for human evaluation due to its unique position as the earliest and smallest parametric model among the LLMs tested. Its smaller size increases the likelihood of it producing a wider range of reasoning and contextual errors, making it an ideal candidate for error analysis. By starting with a model that has fewer parameters and is more prone to subtle reasoning gaps, we can thoroughly evaluate and better understand the types of errors that may occur. This approach maximized the chances of capturing diverse error types that could be missed in larger, more sophisticated models, which tend to exhibit fewer surface-level mistakes. Table 7-11 contain few examples of human analysis of complexity of legal reasoning and error annotations performed for the reasoning chains. The text written in \textcolor{blue}{blue} font in the tables represents the human analyses and annotations performed.

\section{The LLM-based `Auto-Evaluator'}
\label{section:auto-eval}

\paragraph{Rationale behind using LLM-based Auto-Evaluator:} The primary rationale for employing an LLM-based auto-evaluator is its scalability and efficiency compared to human annotation. While human evaluators required approximately 30–60 minutes per reasoning chain to accurately identify and categorize errors in the \textit{`Civ. Pro.'} dataset, the auto-evaluator can process large volumes of reasoning chains significantly faster and at a lower cost. This efficiency becomes even more critical in real-world legal contexts, where expert reviews demand extensive time and resources. Modern LLMs, trained on vast amounts of data (including legal texts), demonstrate state-of-the-art performance in natural language understanding and reasoning. Although our results and analyses show that LLM-based reasoning systems are not yet fully error-free, the rapid improvements in LLM reasoning capabilities and the decreasing costs of inference make them a promising solution for scalable, cost-effective error detection.

\paragraph{Implementation:}At the premise level, four separate prompt-based evaluation pipelines (LLM-based error detectors) have been implemented. Three of the pipelines utilize a single call to GPT-4o, where the information provided to the LLM is broken into three parts: a. The system prompt which contains the information about the error taxa \textbf{(the knowledge base)} b. \textbf{important instructions} provided to the LLM evaluator on how to evaluate and detect errors and c. We also provide an \textbf{in-context learning example} of how the human annotation was carried out for the model to follow and replicate the annotation format. 

\begin{figure*}
    \centering     \includegraphics[width=1.0\linewidth]{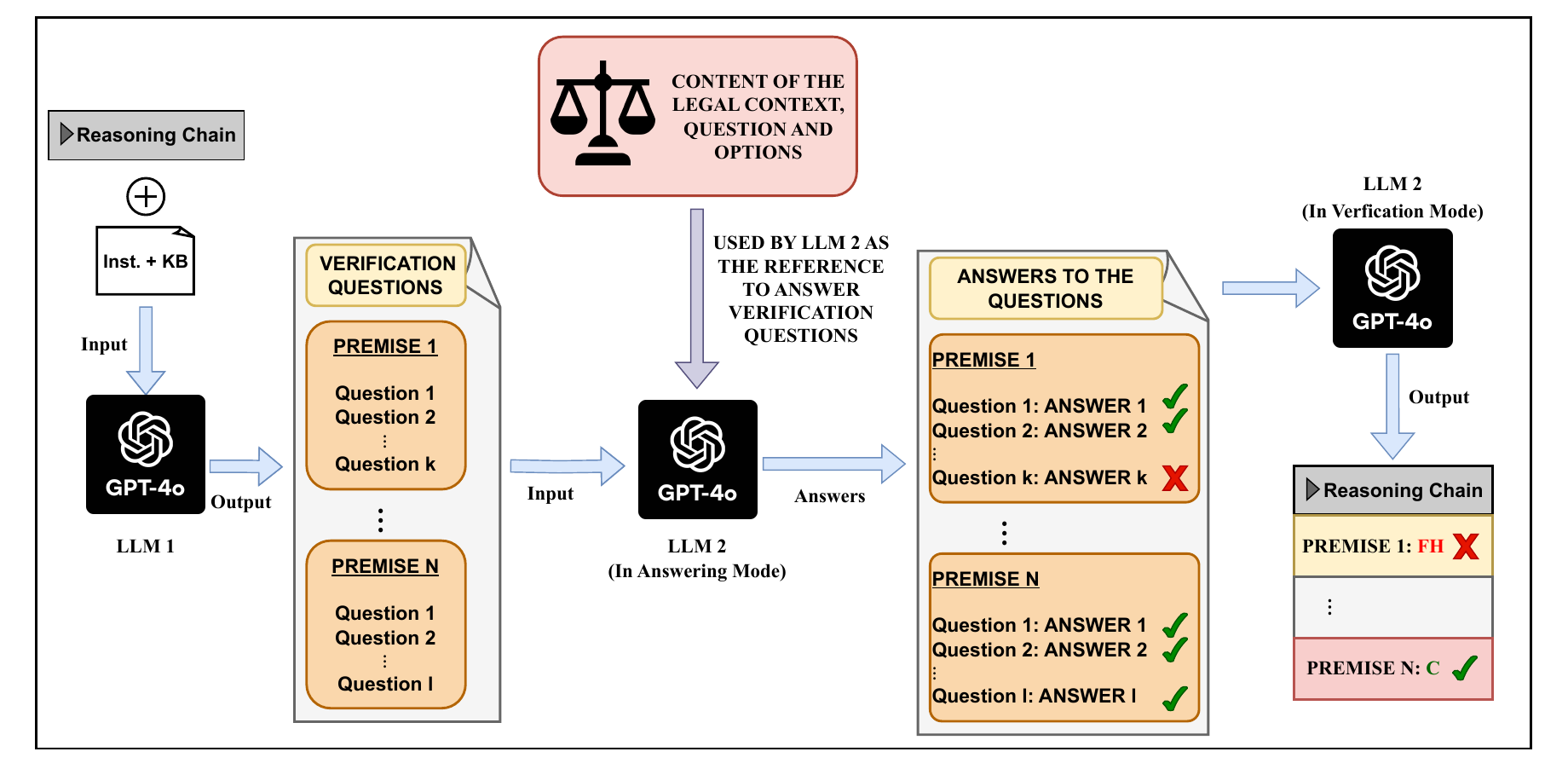}
    \caption{Schematic representation of the `Multi-Call' LLM-based auto-evaluator for 'Factual-Hallucination' error detection. `Inst + KB' includes instructions for error analysis and the knowledge base with error definitions, based on the proposed error taxonomy. The \textcolor{green}{green} check-mark indicates the verification question being correctly answered and the \textcolor{red}{red} cross-mark represents the presence of a contradiction in the content of the premise with the provided context.}
    \label{fig:autoeval-multi-call}
\end{figure*}

\begin{figure*}[t]
    \centering     
    \includegraphics[width=0.965\linewidth]{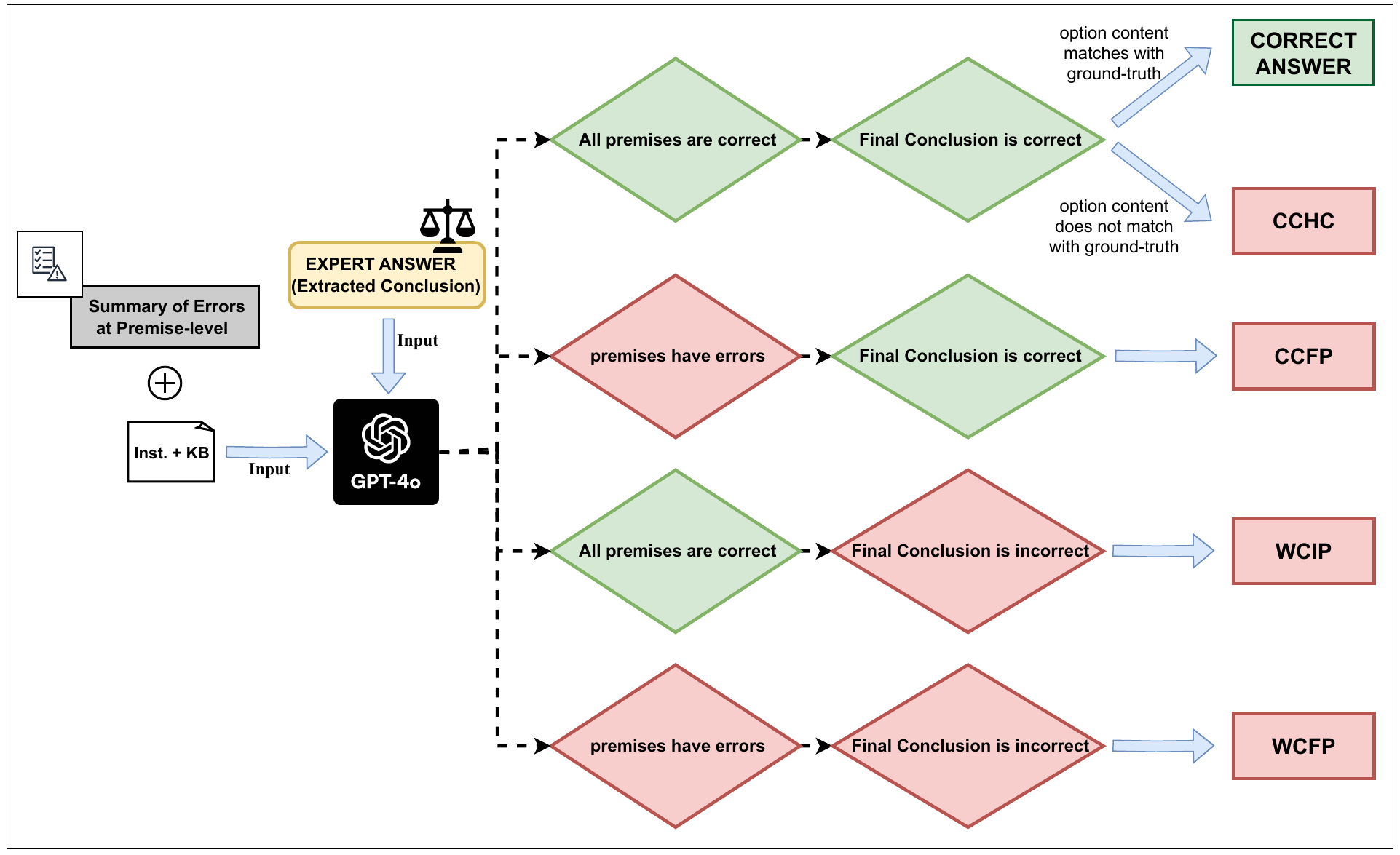}
    \caption{Schematic representation of the  LLM-based auto-evaluator system for error detection at conclusion-level. The GPT-4o LLM represents the `LLM evaluator 5' in Figure \ref{fig:autoeval-system}. The dotted lines represent the conditional paths of which only one will be true and lead to respectively conclusion-level error being labelled to the conclusion, Here, `CCHC' represents `Correct conclusion with Hallucinated Content, `CCFP' represents `Correct Conclusion from False Premises', `WCIP' represents `Wrong Conclusion from Incomplete Premises' and `WCFP' represents `Wrong Conclusion from False Premises'.} 
    \label{fig:autoeval-conclusion-level}
\end{figure*}

\begin{figure*}[t]
    \centering     
    \includegraphics[width=1.0\linewidth]{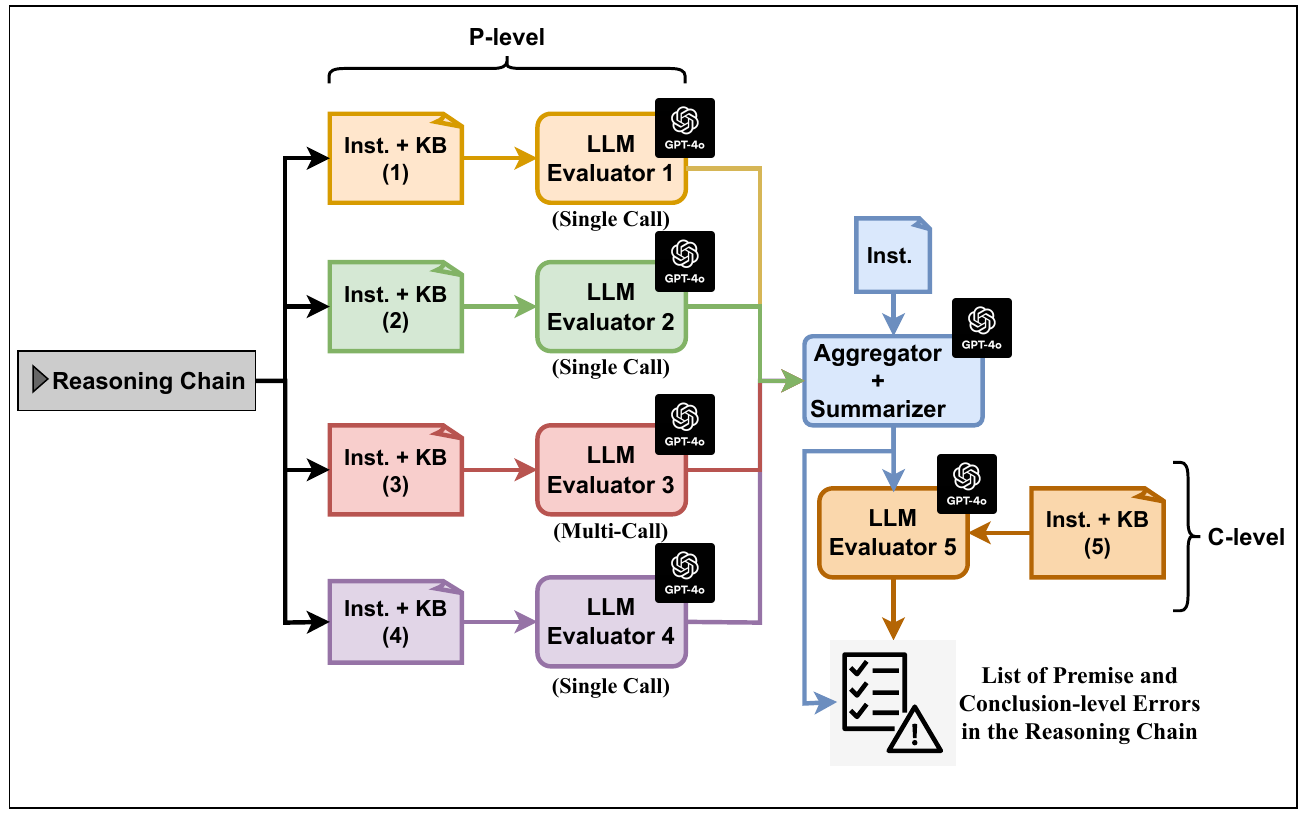}
    \caption{Schematic representation of the  LLM-based auto-evaluator system for error detection. `Inst + KB' with their respective numbers in brackets includes instructions for error analysis and the knowledge base with error definitions, based on the proposed error taxonomy. An example prompt structure (for LLM evaluator 2) has been shown in Figure \ref{fig:annotation-guideline}. Here, 'P-level' denotes the Premise-level, and `C-level' denotes the Conclusion-level. The schematics of the `Single-Call' and `Multi-call' GPT-4o LLM-based evaluators are represented in Figure 
 \ref{fig:autoeval-single-call} and \ref{fig:autoeval-multi-call} respectively.} 
    \label{fig:autoeval-system}
\end{figure*}

The first pipeline (Figure \ref{fig:autoeval-system}) is designed to simply detect whether a premise contains an error, without assigning a specific label from the error taxonomy. The second pipeline detects whether a premise contains a `Misinterpretation' error. The third pipeline does the same for detecting an `Irrelevant Premise'. Along with this, the input prompt includes the step-by-step reasoning chain. The LLM is expected to classify the error and provide an explanation for the classification. The fourth pipeline \ref{fig:autoeval-multi-call}, which has been designed to detect 'Factual Hallucination' errors, consists of a `Multi-call' LLM system where primarily two calls to GPT-4o are made: 1. The first call to GPT-4o is used to create fact-verification questions about various aspects of a particular premise. 2. The second call to GPT-4o is used to answer these verification questions by referencing the legal context and content of the questions and the options. A premise is classified as containing factual hallucination(s) if the answers to its verification questions reveal a contradiction between its content and that of the provided input context. The premise-wise evaluation results of four pipelines are aggregated and summarized by a final `summarization' call to GPT-4o which summarizes and enumerates all possible errors detected in a single premise. The aggregated and summarized results are then sent to the conclusion-level error evaluator which performs a conditional mapping procedure (as shown in Figure \ref{fig:autoeval-conclusion-level} to assign the conclusion-level errors.

\begin{figure}[H]
    \centering     \includegraphics[width=0.90\linewidth]{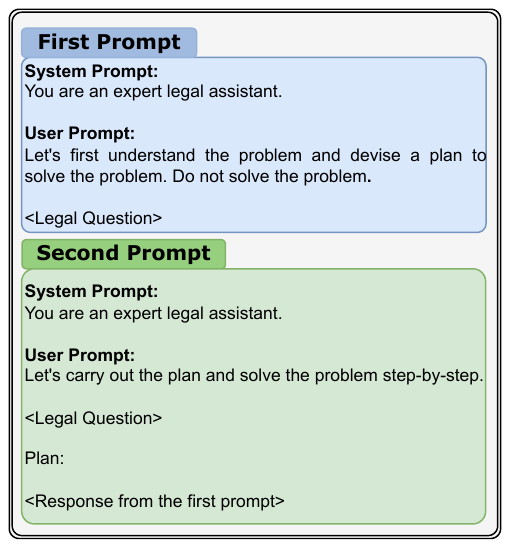}
    \caption{Prompt used to implement 'Plan-and-Solve' technique.} 
    \label{fig:plan-and-solve}
\end{figure}


GPT-4o (Refer Figure \ref{fig:autoeval-system}) was used to aggregate and summarize all the errors detected by the separate premise-level error detectors and summarize them for each premise separately. This had a corrective effect as the most appropriate errors (or combination of multiple errors) were summarized for each premise. An example of this corrective effect is shown in Figure \ref{fig:auto-eval-annot-example}, where both the `Misinterpretation' and `Factual Hallucination' auto-evaluators flag a premise with their respective labels and explanations. However, the summarizer LLM correctly identifies `Misinterpretation' as the most accurate error classification for that premise. The LLM also filtered out unnecessary text, retaining only the premises that contained errors. This process ensured that only the premise steps flagged with errors by the auto-evaluator were forwarded as input to the conclusion-level error analyzer. By focusing on the erroneous premises, this approach streamlined the error analysis process, enabling more efficient and targeted evaluation of how these errors impact the final conclusion.

\begin{figure}[H]
    \centering     \includegraphics[width=0.97\linewidth]{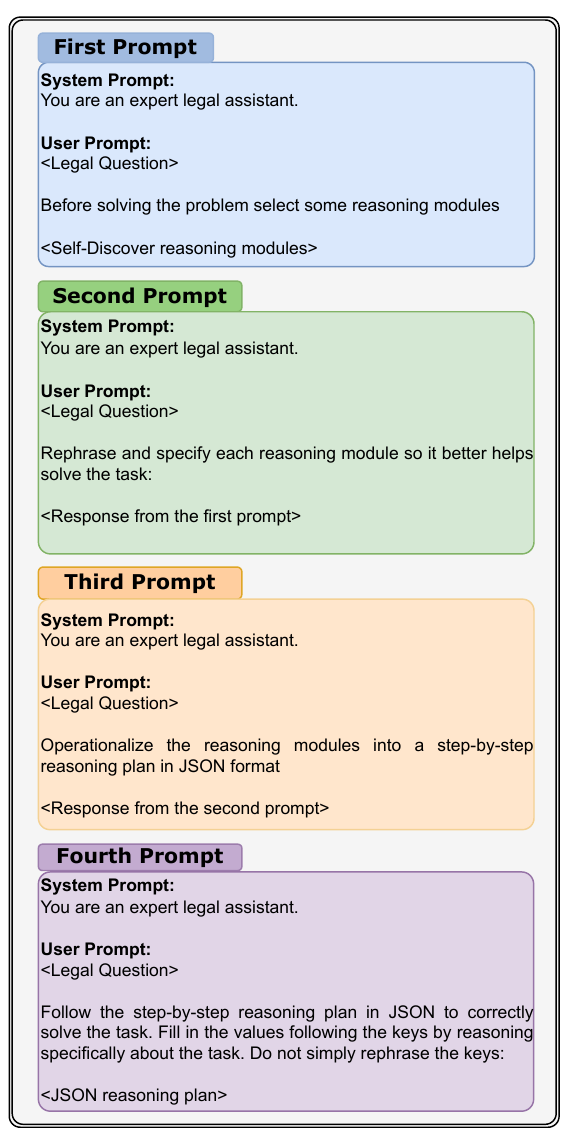}
    \caption{Prompt used to implement `Self-Discovery' technique.} 
    \label{fig:self-discovery}
\end{figure}

\section{Auto-Evaluator Effectiveness}
\label{section:auto-eval-agree}
Based on the errors found in the reasoning chains by human evaluators (as shown in Tables \ref{table:manual-evaluation-premise} and \ref{table:manual-evaluation-conclusion}), the effectiveness of GPT-4o-based auto-evaluator was measured using agreement of step-level presence or absence of errors at both the premise and conclusion level (refer Tables \ref{table:auto-eval-agreement-premise} and \ref{table:auto-eval-agreement-conclusion}). Agreement occurs only when both correct error category and matching error description is generated by the auto-evaluator through semantic error explanation match (refer \textsection \ref{section:llm_aided_eval}). Recall metric was chosen as the appropriate measure for agreement as it is crucial to identify if the auto-evaluator correctly identified an error in a premise or conclusion level step. The recall percentage of detecting an error at the premise level across four LLMs ranged from 86.17$\%$ to 93.85$\%$. The recall percentage range for detecting an error-free premise step was from 83.87$\%$ to 90.6\% (refer Table \ref{table:auto-eval-agreement-premise}). Similarly, at conclusion level, the average recall percentages ranges on agreement on the presence and absence of errors are mentioned in Table \ref{table:auto-eval-agreement-conclusion}. The current auto-evaluator system cannot detect 'Correct Conclusion from Incomplete Premise(s)' (CCIP), as it cannot distinguish it from a `Correct Conclusion from Correct Premises' scenario. While this is a drawback, the `CCIP' error is very rare in its occurrences as an error category. 

\begin{figure}[h]
    \centering     \includegraphics[width=0.97\linewidth]{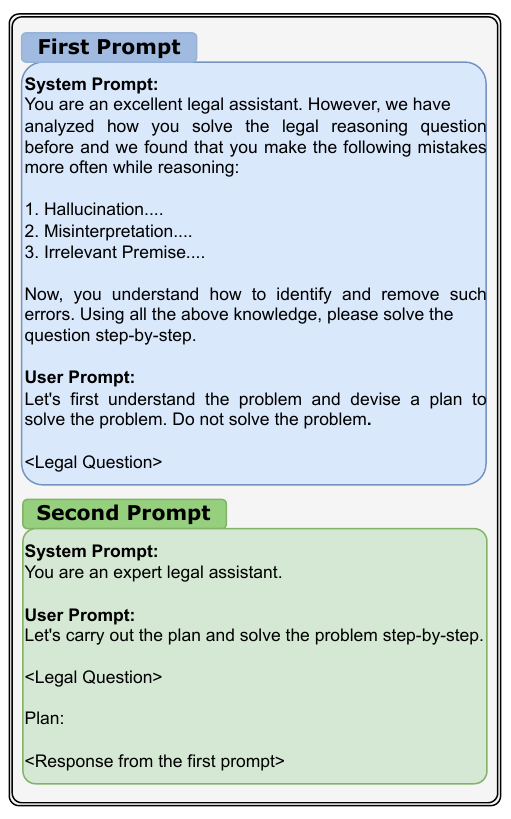}
    \caption{Prompt used to implement `Plan-and-Solve with error feedback' technique. Error feedback can be added to other prompting strategies in the same way.} 
    \label{fig:plan-and-solve-with-feedback}
\end{figure}

\begin{figure}[h]
    \centering     \includegraphics[width=0.90\linewidth]{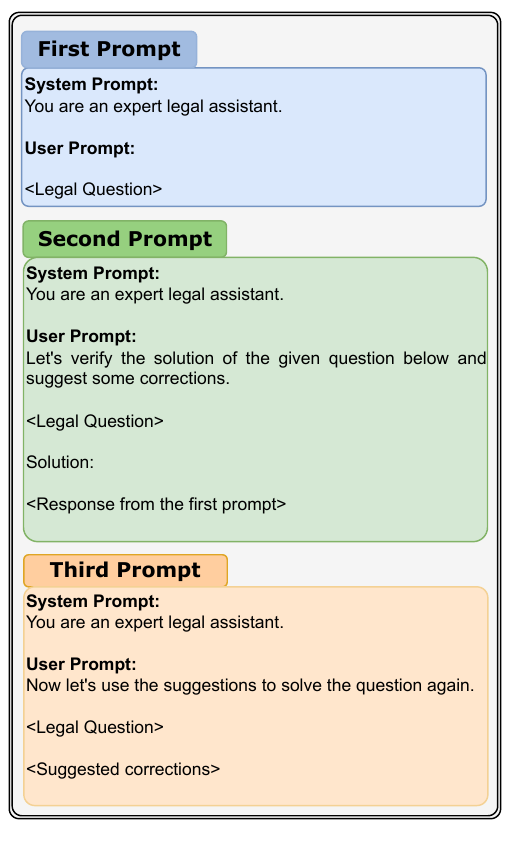}
    \caption{Prompt used to implement `Self-Correct' technique.} 
    \label{fig:self-correct}
\end{figure}
\section{The `Aggregator + Summarizer' LLM}
\label{section: summ_agg_llm}

\begin{table*}[!htbp]
\small
\centering
\begin{tabular}{lcc}
\toprule
LLM   & Statistics & Agreement \% (Recall) \\ 
\midrule
        Mistral-7B-v2-Instruct (135) & CC = 81, NC = 13, EE = 35, NE = 6 & R(C) = 81/94 = 86.17\%, R(E) = 35/41 = 85.3\% \\
        Llama-3-8B-Instruct (145) & 	CC = 107, NC = 7, EE = 26, NE = 5 & R(C) = 107/114 = 93.85\%, R(E) = 26/31 = 83.87\% \\
        GPT-3.5-turbo (109) & 	CC = 69, NC = 8, EE = 29, NE = 3 & R(C) = 69/77 = 89.61\%, R(E) = 29/32 = 90.6\% \\
        GPT-4-turbo (148) & CC = 104, NC = 16, EE = 25, NE = 3 & R(C) = 104/120 = 86.66\%, R(E) = 25/28 = 89.2\% \\
\bottomrule
\end{tabular}
\caption{Agreement statistics between the GPT-4o-based auto-evaluator and human evaluators for the same 30 reasoning chains at the premise level across four LLMs (refer to Table \ref{table:manual-evaluation-premise}). Numbers inside parentheses denote the total number of premise-level reasoning steps evaluated. `CC' represents agreement between the auto-evaluator and human evaluators on the absence of an error in a reasoning step, while 'NC' denotes disagreement on the absence of an error. `EE' indicates agreement on the presence of an error, and 'NE' denotes disagreement on the presence of an error. `R(C)' refers to the recall percentage for agreement on error-free steps between the auto-evaluator and human evaluators, whereas `R(E)' denotes the recall percentage for agreement on steps containing errors.}
\label{table:auto-eval-agreement-premise}
\end{table*}
\begin{table*}[!htbp]
\small
\centering
\begin{tabular}{lcc}
\toprule
LLM   & Statistics & Agreement \% (Recall) \\ 
\midrule
        Mistral-7B-v2-Instruct (30) & CC = 3, NC = 1, EE = 22, NE = 4 & R(C) = 3/4= 75\%, R(E) = 22/26 = 84.61\% \\
        Llama-3-8B-Instruct (30) & 	CC = 9, NC = 0, EE = 20, NE = 1 & R(C) = 9/9 = 100\%, R(E) = 20/21 = 95.23\% \\
        GPT-3.5-turbo (30) & 	CC = 4, NC = 3, EE = 21, NE = 1 & R(C) = 4/7 = 57\%, R(E) = 21/22 = 95.45\% \\
        GPT-4-turbo (30) & 	CC = 12, NC = 4, EE = 13, NE = 1 & R(C) = 12/16=75\%, R(E) = 13/14 = 92.85\% \\
\bottomrule
\end{tabular}
\caption{Agreement statistics between the GPT-4o-based auto-evaluator and human evaluators for the same 30 reasoning chains at the conclusion level across four LLMs (refer to Table \ref{table:manual-evaluation-conclusion}). `CC' represents agreement between the auto-evaluator and human evaluators on the absence of an error in a conclusion, while `NC' denotes disagreement on the absence of an error. `EE' indicates agreement on the presence of an error, and `NE' denotes disagreement on the presence of an error. `R(C)' refers to the recall percentage for agreement on error-free conclusions between the auto-evaluator and human evaluators, whereas `R(E)' denotes the recall percentage for agreement on conclusions containing errors.}
\label{table:auto-eval-agreement-conclusion}
\end{table*}

Furthermore, we conducted another experiment, replacing GPT-4o with Gemini-1.5-Flash as the backbone LLM for the auto-evaluator system. The prompts provided to auto-evaluator system were unchanged. The recall rate of identifying an error in the reasoning step at the premise-level for Gemini-based auto-evaluator on the Mistral-7B-v2-Instruct generated reasoning chains, when compared with human annotations, was found out to be $\sim$78.1\%. The recall-rate for identifying correct premise-level reasoning steps highly decreased to $\sim$20.61\%. This indicates that Gemini-1.5-Flash, while comparable to GPT-4o in identifying errors, was less effective overall as an auto-evaluator due to a higher number of false-negative predictions (with larger number of error-free steps were incorrectly identified as containing errors). Although fine-tuning and testing a dedicated LLM for error detection was not conducted in this study, the authors acknowledge it as an interesting avenue for future improvements and research directions.

\section{Prompting-techniques for Error Mitigation}
\label{section:prompting_strategies}
\begin{figure*}[t]
    \small
    \centering     \includegraphics[width=0.96\linewidth]{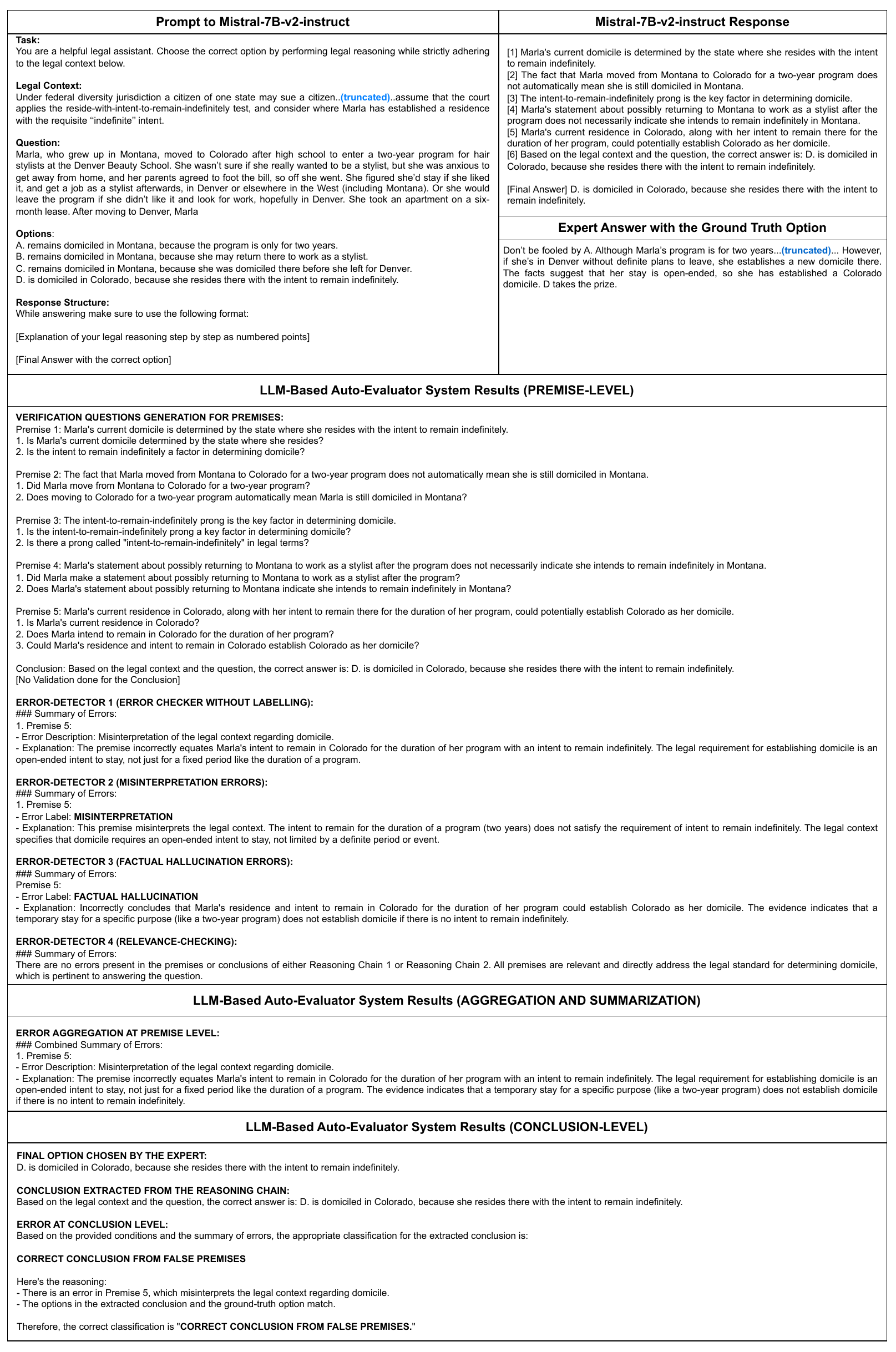}
    \caption{Example LLM Auto-Evaluator Response: Top: Zero-shot CoT prompt and LLM response. Bottom: Outputs from the LLM-based auto-evaluator.} 
    \label{fig:auto-eval-annot-example}
\end{figure*}

We carried out several experiments on the legal reasoning dataset, employing widely used prompting techniques alongside the most frequently observed errors we found through \textsection \ref{section:error_taxonomy} with the aim to explore the possibility of enhancing the reasoning capabilities of both closed-source and open-source LLMs. Four prompting techniques were utilized: (1) Chain-of-Thought \citep{wei2022chainofthought} (2) Plan-and-Solve \citep{wang2023plan} (3) Self-Correct  \citep{zhang2024smalllanguagemodelsneed}, and (4) Self-Discovery  \citep{zhou2024selfdiscoverlargelanguagemodels}. We used the zero-shot CoT (Figure \ref{fig:auto-eval-annot-example}) method as the baseline method in which the LLM is prompted to provide the final answer along with step-by-step reasoning. 

Plan-and-Solve prompts the LLM first to generate a plan to solve the problem without solving it and after that the LLM carries out the self-suggested plan to get the final answer as shown in Figure \ref{fig:plan-and-solve}. Self-Correct uses self-verification and self-refining to improve the reasoning ability of the LLMs as shown in Figure \ref{fig:self-correct}. Self-Discover as shown in Figure \ref{fig:self-discovery} utilizes self-discover reasoning modules to create an explicit reasoning structure to follow to solve the problem. For running the above prompting strategies with error taxonomy as the feedback, we include a detailed description of the error taxonomy in the system prompt. For instance, Plan-and-Solve with error feedbacks is shown in Figure \ref{fig:plan-and-solve-with-feedback}.

\begin{table}[!htbp]
\small
\centering
\begin{tabular}{lcc}
\toprule
Model & Zero-shot($\uparrow$) & Few-shot($\uparrow$)\\ 
\midrule
Llama-3.1-8B-Instruct & 53.71  & 52.87 \\
Llama-3.1-70B-Instruct & 76.74  & 71.51 \\
GPT-4o & 82.56  & 80.23 \\
\bottomrule
\end{tabular}
\caption{The zero-shot and few-shot results of LLMs in CoT setting.}
\label{table:few-shot}
\end{table}

\section{Few-shot prompting on performance of LLMs}
\label{section:few-shot}

We evaluated the performance of several large language models (LLMs) on a set of legal reasoning questions using zero-shot and few-shot prompting, as well as chain-of-thought (CoT) prompting. For Llama models, we used one example and for other we used 3 examples in few-shot training. Our results indicate that LLMs perform best in the zero-shot setting, where no in-context examples are provided. We hypothesize that the diverse nature of legal reasoning questions limit the effectiveness of in-context learning, as it restricts the model's ability to generalize beyond the provided examples.



\begin{figure*}[tp]
    \centering
    
    \begin{subfigure}{0.32\textwidth}
        \centering
        \includegraphics[width=\textwidth]{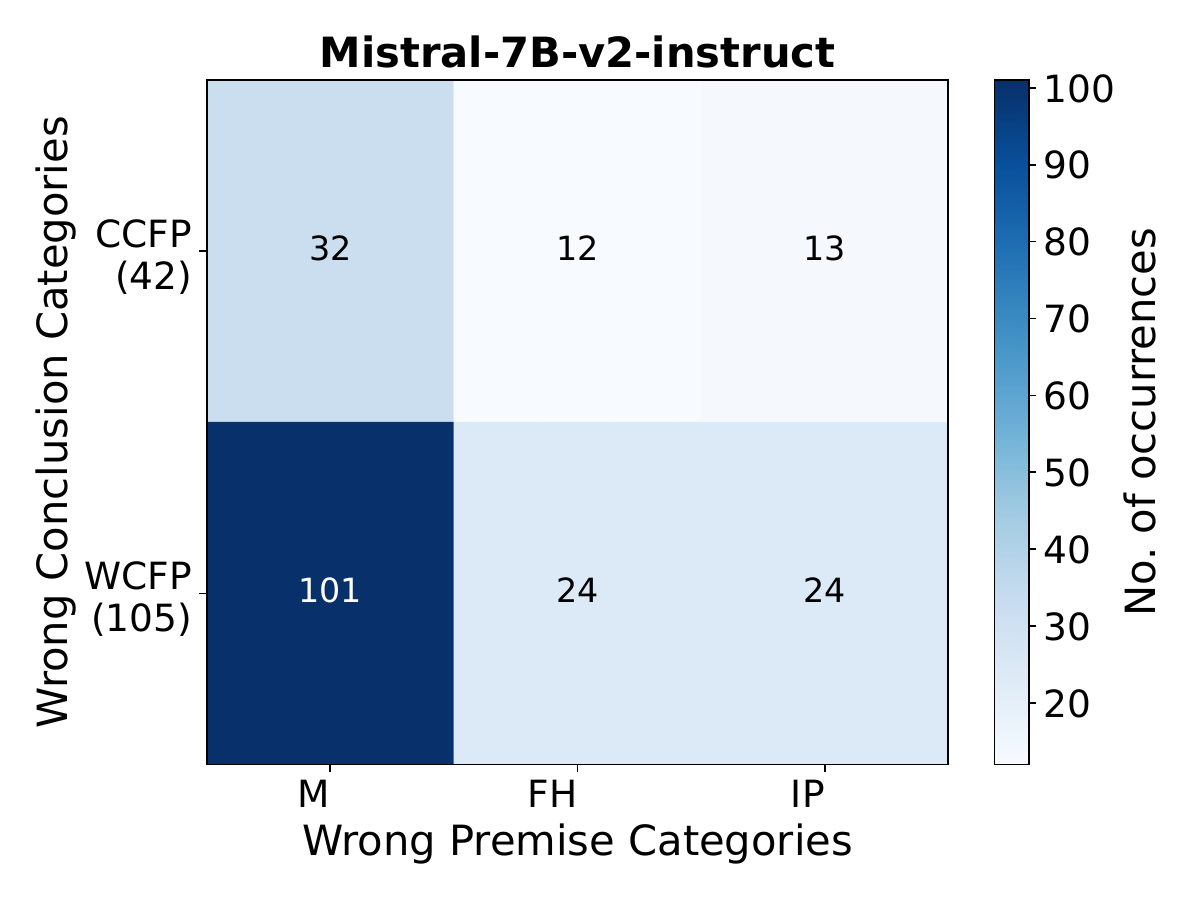}
    \end{subfigure}
    \hfill 
    \begin{subfigure}{0.32\textwidth}
        \centering
        \includegraphics[width=\textwidth]{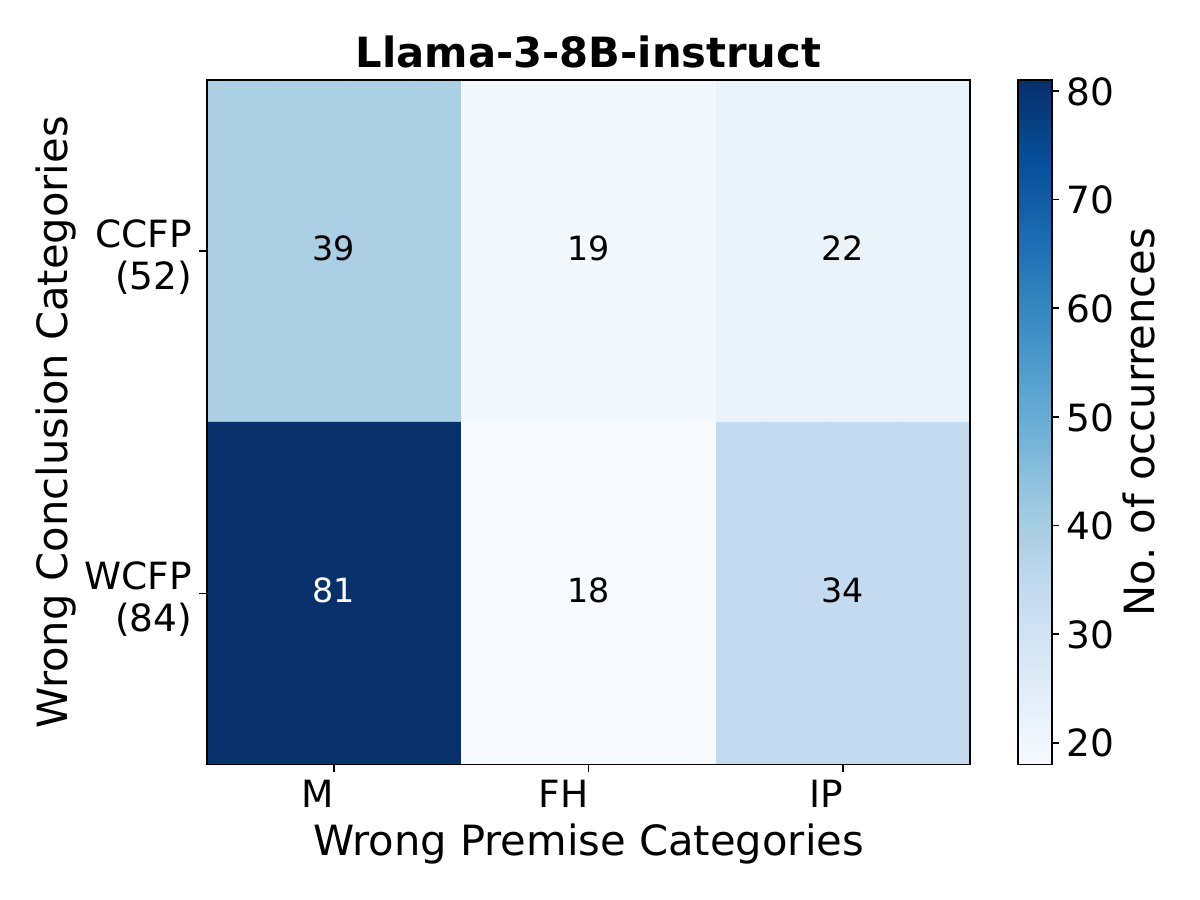}
    \end{subfigure}
    \hfill
    \begin{subfigure}{0.32\textwidth}
        \centering
        \includegraphics[width=\textwidth]{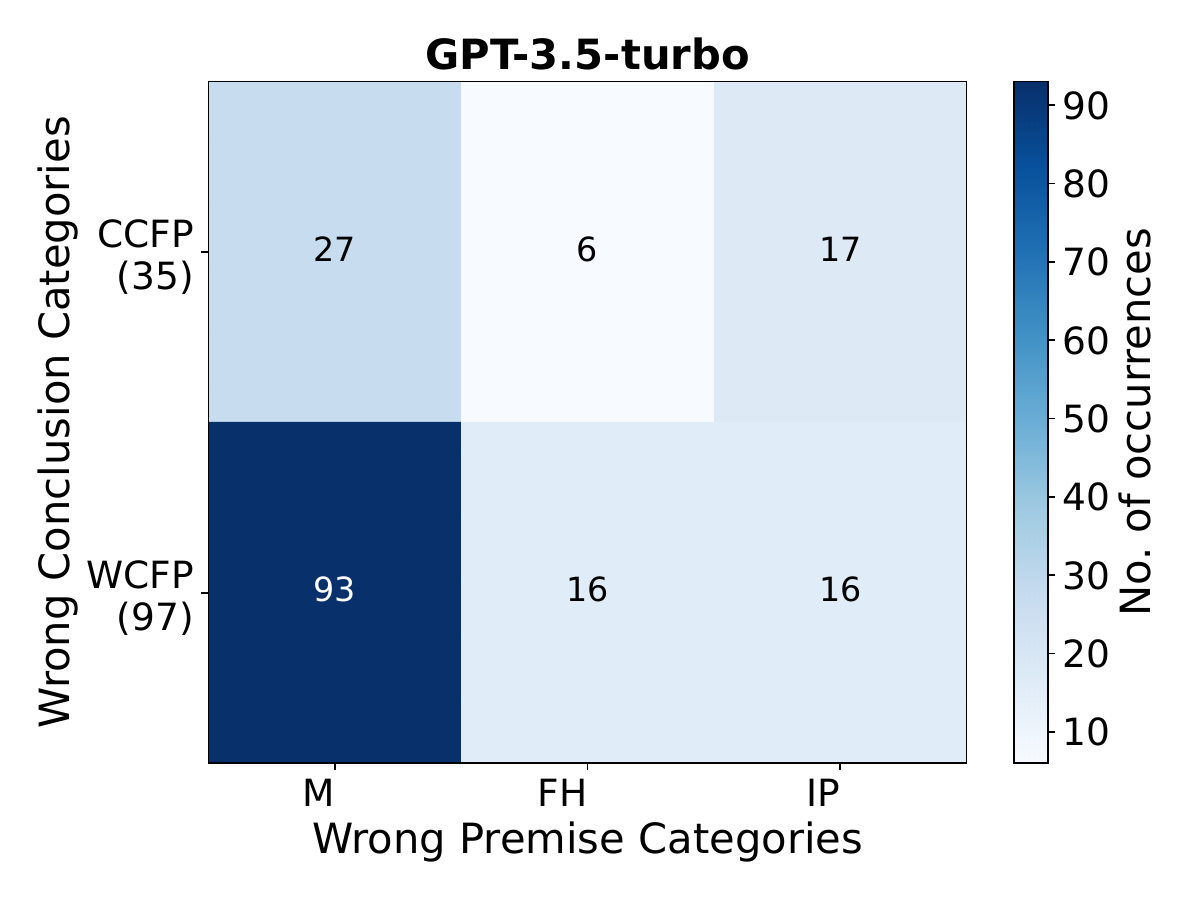}
    \end{subfigure}
    
    \vspace{1em} 

    \begin{subfigure}{0.32\textwidth}
        \centering
        \includegraphics[width=\textwidth]{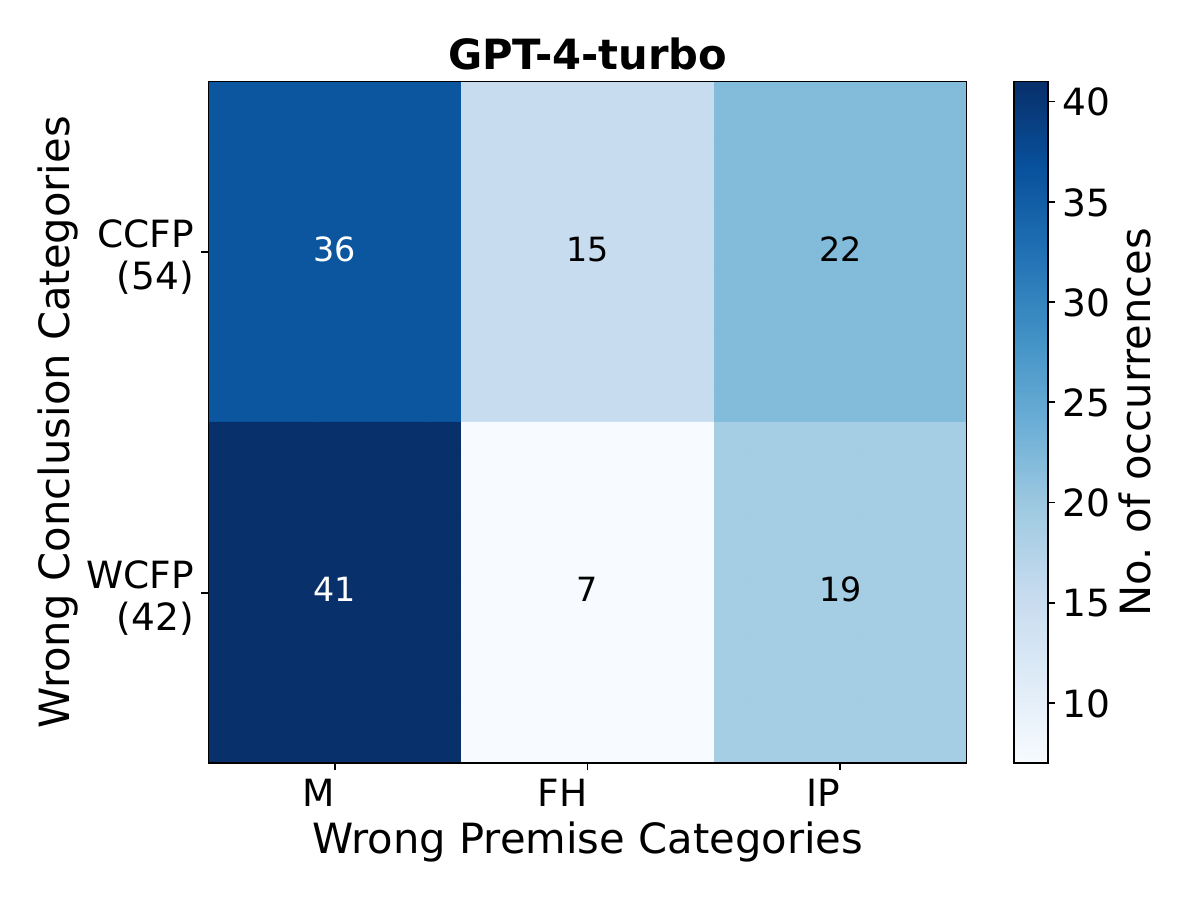}
    \end{subfigure}
    \hfill
    \begin{subfigure}{0.32\textwidth}
        \centering
        \includegraphics[width=\textwidth]{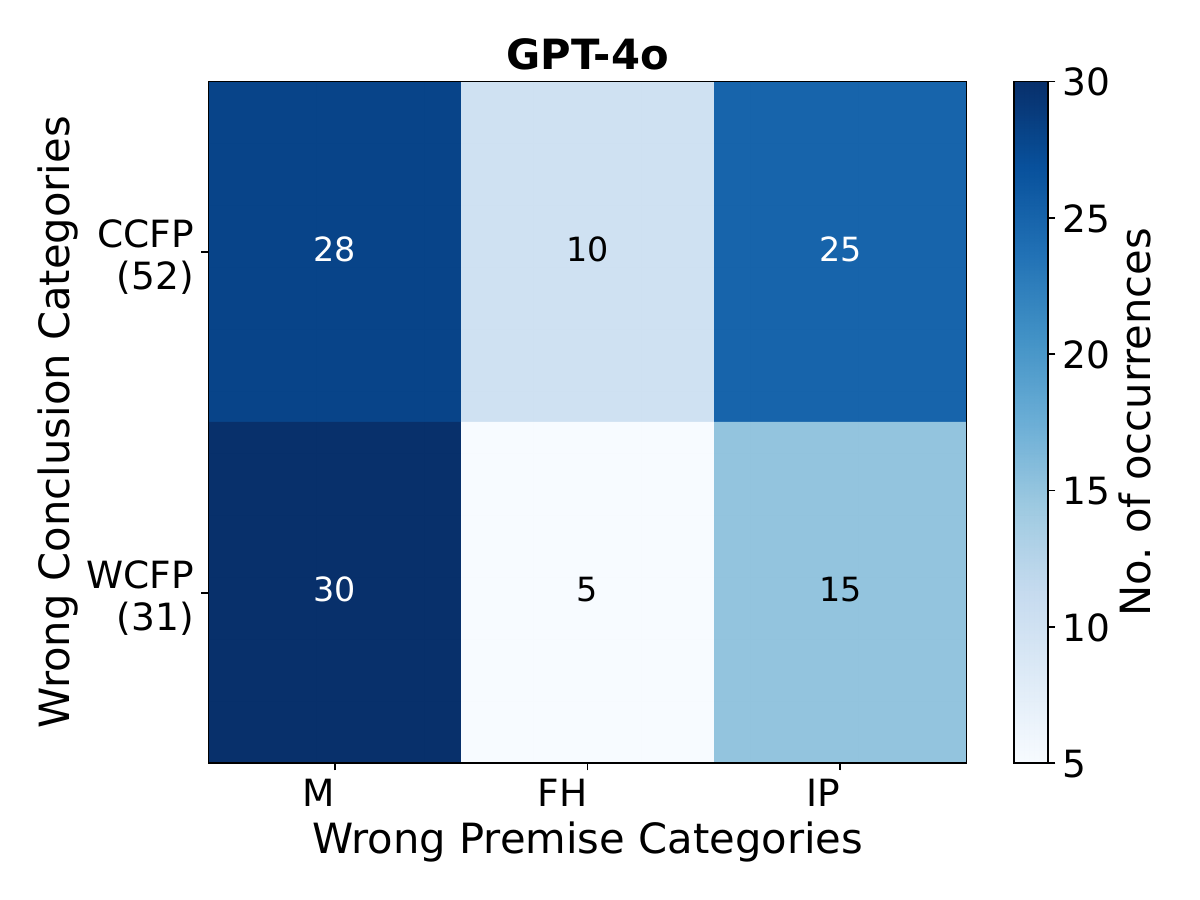}
    \end{subfigure}
    \hfill
    \begin{subfigure}{0.32\textwidth}
        \centering
        \includegraphics[width=\textwidth]{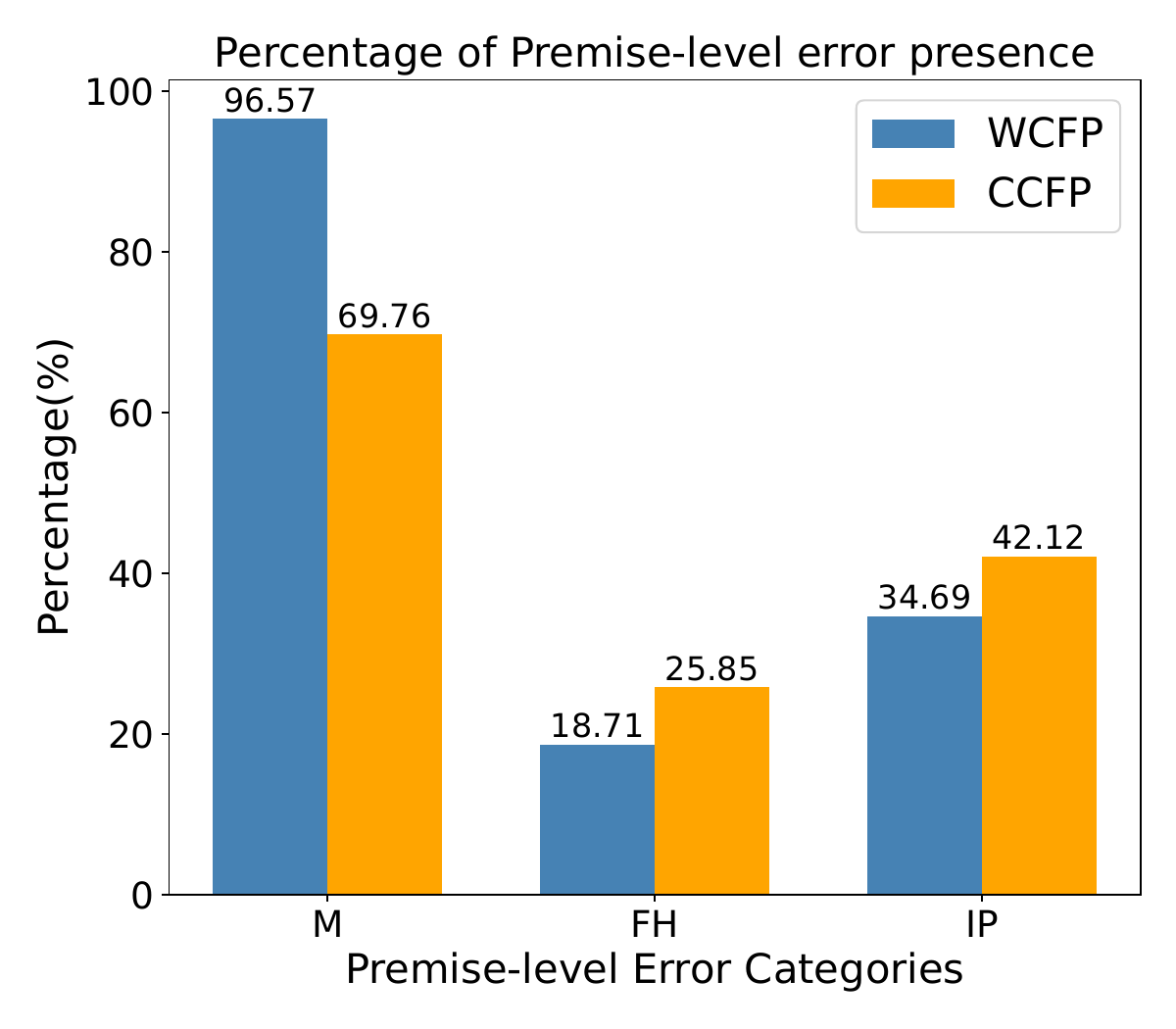}
    \end{subfigure}
    \caption{The first five sub-figures in the preceding section display the error distribution for premise-level errors in instances where the conclusion contains an error, across five different LLMs. The dimension in the x-axis represents the categories of errors at the premise level. The dimension in the y-axis represents the errors at the conclusion level. Here `M' denotes `Misinterpretation', `FH' denotes Factual Hallucination, `IP' denotes Irrelevant Premise, `WCFP' denotes Wrong Conclusion from False Premise(s), and `CCFP' denotes Correct Conclusion from False Premises. Here the counts in the heatmap represent the presence of one (or more) premise-level error(s) (in the x-axis) in a reasoning chain containing the specified conclusion-level error (in the y-axis)}
    \label{fig:prem_conc_corr}
\end{figure*}

\begin{figure*}[ht]
    \centering     
    \includegraphics[width=0.6\linewidth]{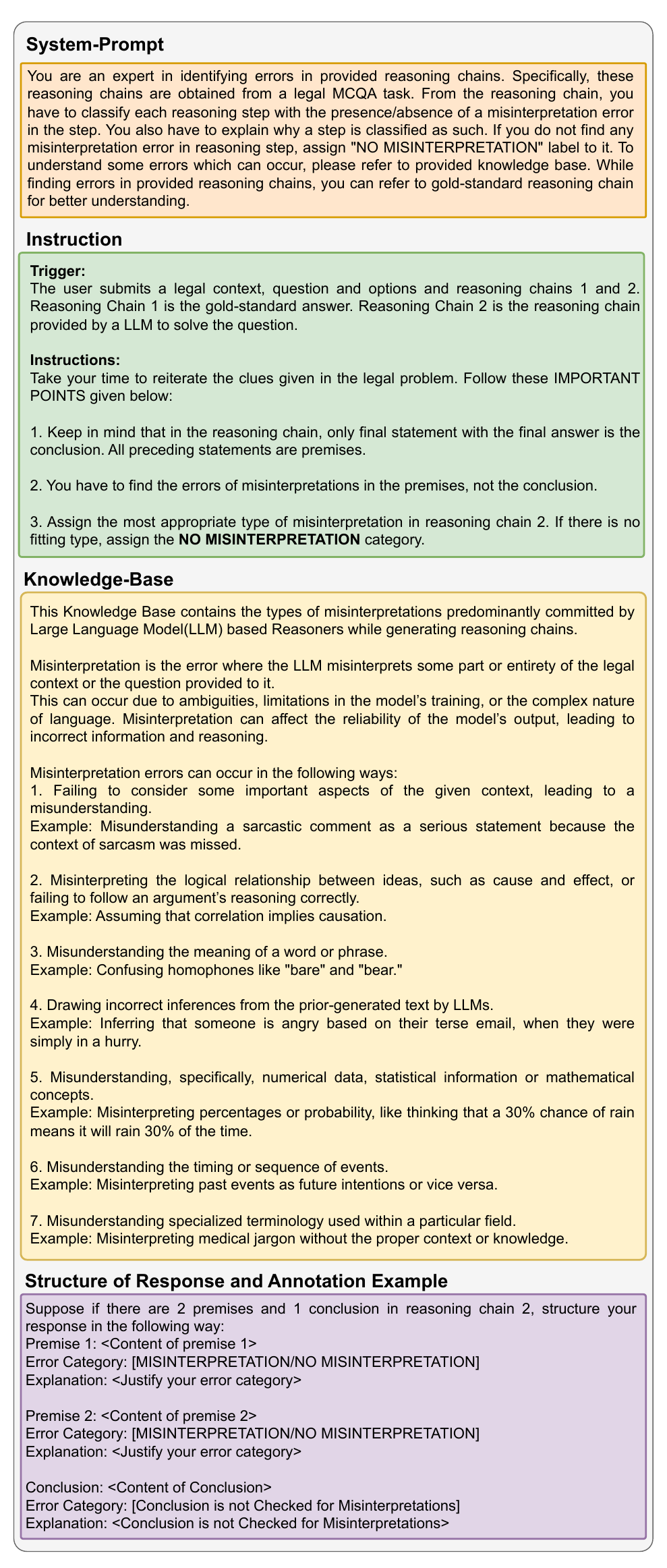}
    \caption{Prompt structure for LLM evaluator 2 from Figure \ref{fig:autoeval-system}. Appropriate Instructions and Knowledge Base for detecting `Misinterpretation' errors are provided in the prompt. Similar prompt structures and appropriate knowledge bases have been provided to all other LLM-based evaluators.} 
    \label{fig:misinterpretation-prompt}
\end{figure*}

\clearpage

\begin{figure*}[p]
  \centering
  \begin{subfigure}{\linewidth}
    \centering
    \includegraphics[width=\linewidth]{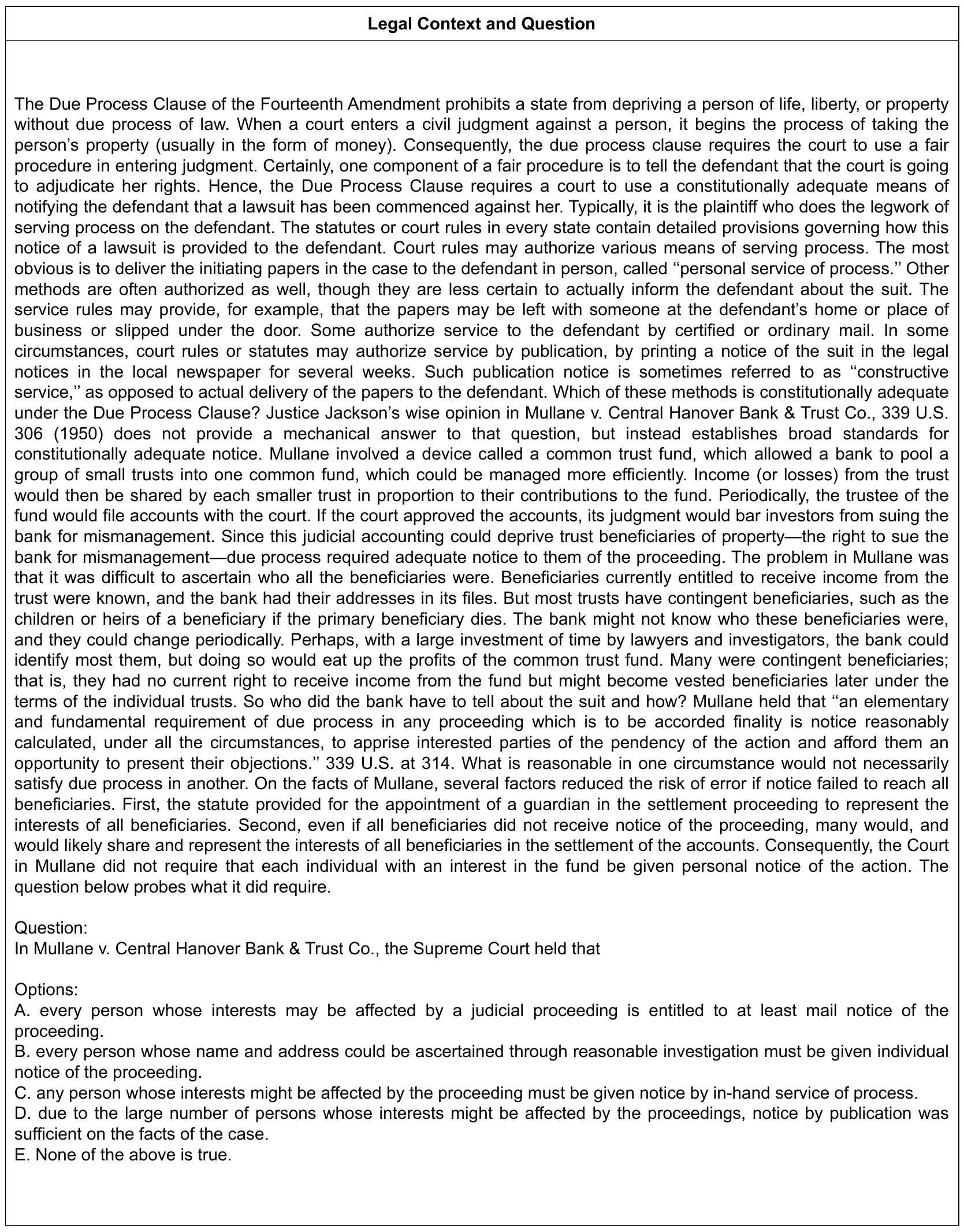}
    \caption{Part 1: Contains the Legal Context, Questions and Options.}
    \label{fig:subfig1}
  \end{subfigure}
  \caption{A Legal scenario which showcases the complexity of legal reasoning involved. This is one of the cases where all the LLMs under evaluation provided wrong options as answer.}
  \label{fig:legal-reasoning-complexity-part-1}
\end{figure*}

\clearpage

\begin{figure*}[p]\ContinuedFloat
  \centering
  \begin{subfigure}{\linewidth}
    \centering
    \includegraphics[width=\linewidth]{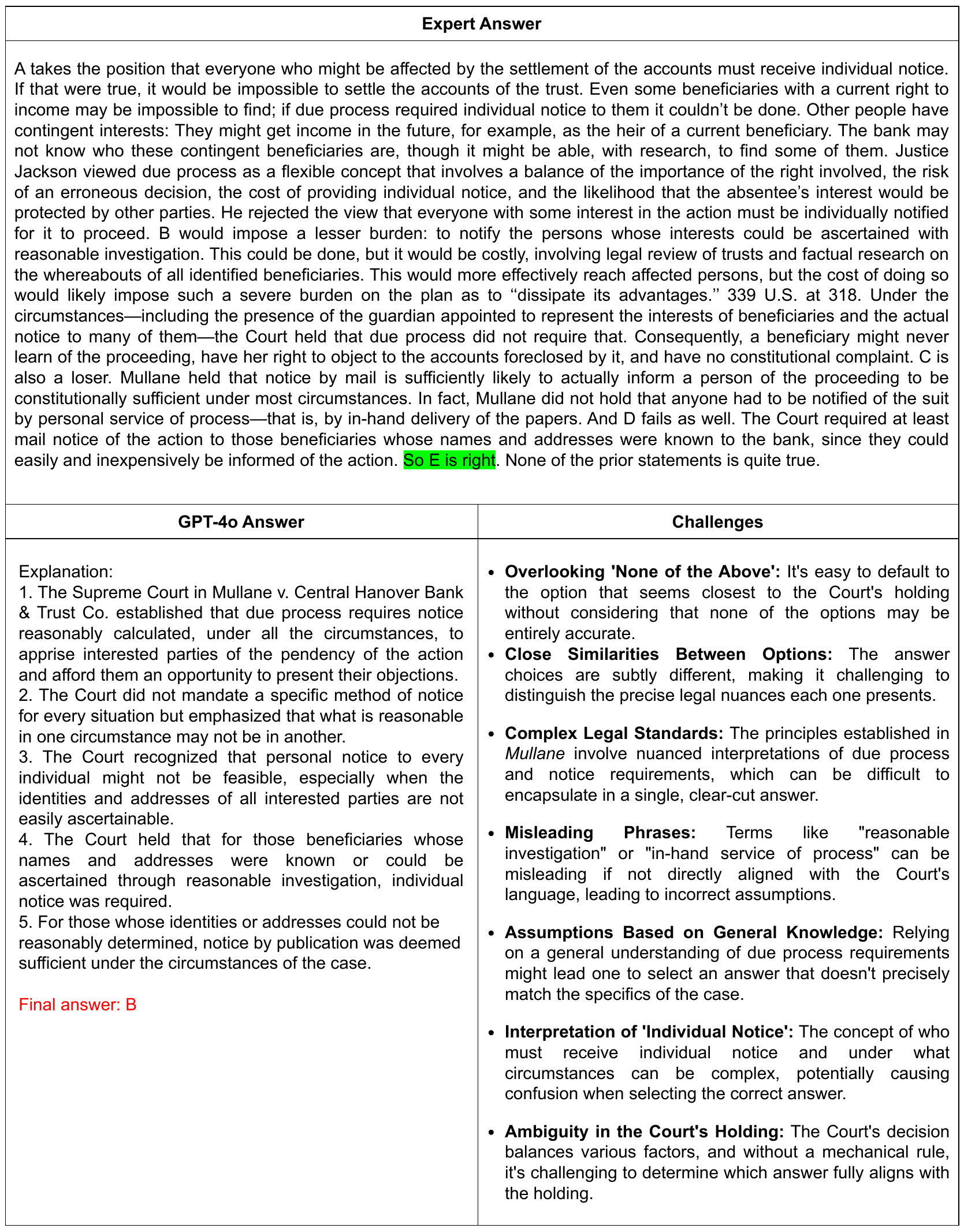}
    \caption{Part 2: Contains the Expert Answer, LLM step-by-step response and Reasoning challenges.}
    \label{fig:subfig2}
  \end{subfigure}
  \caption{Continuation from the previous sub-part of Figure \ref{fig:legal-reasoning-complexity-part-1}. This sub-part contains the expert-answer to a legal scenario, a wrong response provided by GPT-4o and the challenges faced by a legal reasoner to solve this scenario}
  \label{fig:legal-reasoning-complexity-part-2}
\end{figure*}

\begin{table*}[h]
\footnotesize
\centering
\begin{tabular}{p{0.9\linewidth}}
\toprule
\textbf{Legal Context:} Under federal diversity jurisdiction a citizen of one state may sue a citizen of another in federal court, even though her claim arises under state law if she has a colorable claim for more than \$75,000. The state citizenship of a person—as opposed to that of a corporation—is determined by her domicile, that is, the most recent state where she has (1) resided with (2) the intent to remain indefinitely. The ‘‘residence’’ requirement is easily satisfied. Staying overnight in a hotel or a tent will establish ‘‘residence’’ in a state. It is the intent-to-remain-indefinitely prong that gives students problems. A person intends to remain indefinitely in a state if she is residing in the state on an open-ended basis, without the intent to leave at a definite time or on the occurrence of a definite event. You don’t have to swear allegiance forever to a state to acquire domicile there; you only need to reside there ‘‘indefinitely,’’ that is, on an open-ended basis. If a party is living in a state without definite plans to leave, the domicile test regards that state as her ‘‘home.’’ She is there, not as a visitor, but as a citizen. She is, psychologically speaking, at home there, rather than passing through. She may choose to move on, as we all may, but at the moment she has no plans to do so. Remember that, until the two prongs coincide in a new state, your old domicile continues, whether you plan to return to that state or not. If Acari, from Hawaii, leaves for a one-year job acting in a play in California, planning to go to New York afterwards, he remains domiciled in Hawaii, even if he swears that he will never return to Hawaii. He hasn’t acquired a domicile in California, because he doesn’t plan to stay there indefinitely. He hasn’t acquired one in New York either, since he doesn’t reside there yet. Domicile doctrine abhors a vacuum, so it holds that Acari keeps his Hawaii domicile until the two prerequisites come together in another state. In analyzing the question below, assume that the court applies the reside-with-intent-to-remain-indefinitely test, and consider where Marla has established a residence with the requisite ‘‘indefinite’’ intent.                      \\ \midrule
\textbf{Question:} Marla, who grew up in Montana, moved to Colorado after high school to enter a two-year program for hair stylists at the Denver Beauty School. She wasn’t sure if she really wanted to be a stylist, but she was anxious to get away from home, and her parents agreed to foot the bill, so off she went. She figured she’d stay if she liked it, and get a job as a stylist afterwards, in Denver or elsewhere in the West (including Montana). Or she would leave the program if she didn’t like it and look for work, hopefully in Denver. She took an apartment on a six-month lease. After moving to Denver, Marla\\ \textbf{Options:}\\ A. remains domiciled in Montana, because the program is only for two years. \\ B. remains domiciled in Montana, because she may return there to work as a stylist. \\ C. remains domiciled in Montana, because she was domiciled there before she left for Denver. \\ D. is domiciled in Colorado, because she resides there with the intent to remain indefinitely.                                \\ \midrule
\textbf{Human Annotator Analysis of LLM generated Reasoning Chain:}\\ {[}1{]} The question asks about Marla's domicile based on her current residence in Denver, Colorado, and her intentions regarding her stay there. \textcolor{blue}{(Correct)}\\ \\ {[}2{]} According to the legal context, a person's domicile is determined by the state where they reside with the intent to remain indefinitely.\textcolor{blue}{(Correct)}\\ \\ {[}3{]} Marla moved to Denver, Colorado, to attend a two-year hair stylist program. She was unsure about her career choice and had plans to leave the program if she didn't like it. She also mentioned that she might look for work in Denver or other western states, including Montana.\textcolor{blue}{(Correct)}\\ \\ {[}4{]} Marla's lease in Denver is for six months, which indicates that she does not have an open-ended intention to remain in Colorado indefinitely. \\ \textcolor{blue}{(\textbf{Misinterpretation - Omission of parts of the provided context while reasoning}) - (Lease in Denver is for six months (A) -> indicates that she does not have an open-ended intention to remain in Colorado indefinitely (B)) - Here B can not be true just because A is true, there are other factors which need to be true for B to be actually true. (A alone is not sufficient to prove that B is true.) (Incomplete/Insufficient cause for an effect)}\\ \\ {[}5{]} The fact that Marla was domiciled in Montana before moving to Denver does not automatically make her domiciled in Montana again. Domicile is determined by the present intent to remain indefinitely in a state, not by past domicile. \textcolor{blue}{(Correct)} \\ \\ {[}6{]} Based on the information provided, Marla does not meet the requirement of residing in Colorado with the intent to remain indefinitely. \textcolor{blue}{(Wrong Conclusion from False Premise) - Wrong Conclusion as a consequence of {[}4{]}; because of point {[}4{]}, the LLM seems to have drawn a wrong conclusion, Error Propagation from {[}4{]} to {[}6{]})} \\ \\ {[}Final Answer{]} A. remains domiciled in Montana.\\ \bottomrule
\end{tabular}
\caption{Example of a Misinterpretation and Wrong Conclusion due to False Premise. The complexities of this scenario are interpreting Marla's intent to remain indefinitely in Colorado by assessing her open-ended plans and lack of definite departure; distinguishing between temporary residence and establishing a new domicile by considering factors like her enrollment in a two-year program and a six-month lease against her willingness to stay in Colorado long-term; and applying domicile principles amid uncertainty.}
\label{table:human_annotation_example_initial}
\end{table*}

\begin{table*}[h]
\footnotesize
\centering
\begin{tabular}{p{0.9\linewidth}}
\toprule
\textbf{Legal Context:} The most fundamental point about arising-under jurisdiction is that many cases that involve federal issues do not ‘‘arise under’’ federal law. This is the essential message of that civil procedure chestnut, Louisville \& Nashville R.R. v. Mottley, 211 U.S. 149 (1908). In Mottley, the plaintiffs sued the railroad for breach of contract, and alleged in their complaint that the railroad had breached the contract because it believed that a federal statute prohibited it from renewing their passes for free travel. In fact, when it answered the complaint, the railroad did rely on the federal statute as their justification for denying the passes. The parties then litigated that federal question and the federal trial court held that the federal statute did not bar renewing the passes. The United States Supreme Court reversed and ordered the case dismissed for lack of subject matter jurisdiction. The Court held that a case only ‘‘arises under’’ federal law, within the meaning of the federal statute granting arising-under jurisdiction, if the plaintiff relies on federal law as the source of her right to relief. That wasn’t true in Mottley: the plaintiffs had sued for breach of contract, a state law claim...\textcolor{blue}{\textbf{...Truncated}}                      \\ \midrule
\textbf{Question:} Consolidated Corporation sues Garces, a former employee, for business libel, a tort claim. The suit is brought in federal court, and alleges that Garces falsely reported to a federal agency that Consolidated was using watered-down concrete on a federal construction project. Garces answers, admitting that he had made a report to the agency, but alleging that his report was protected by the First Amendment to the United States Constitution, which guarantees the right of free speech. He also files a counterclaim against Consolidated, based on the Federal Whistleblower Act, which authorizes damages for anyone who is dismissed or disciplined for reporting fraud on the federal government. After answering the complaint, Garces moves to dismiss for lack of subject matter jurisdiction. (Assume that there is no jurisdiction based on diversity.) The motion should be\\ \textbf{Options:}\\ A. granted, because the court lacks subject matter jurisdiction over the case. \\ B. denied, because the complaint alleges that Garces reported the violations to a federal agency. \\ C. denied, because the case involves a question arising under the Constitution of the United States. \\ D. denied, because Garces’s counterclaim arises under federal law. \\ E. denied, because Garces waived his objection to jurisdiction by answering the complaint and asserting a counterclaim against Consolidated.                                \\ \midrule
\textbf{Human Annotator Analysis of LLM generated Reasoning Chain:}\\ {[}1{]} Understanding the Mottley Rule: The Mottley rule establishes that a federal court has jurisdiction over a case only if the plaintiff's claim itself arises under federal law. This means the plaintiff must base their right to relief on federal law for the case to be heard in federal court. \textcolor{blue}{(Correct)}\\
\\{[}2{]} Analyzing the Plaintiff's Claim: In this scenario, Consolidated Corporation's claim against Garces is for business libel, which is a state tort claim. The claim does not inherently rely on federal law for the basis of the relief sought.\textcolor{blue}{(Correct)}\\
\\{[}3{]} Considering the Defendant's Response: Garces's answer and the assertion of a First Amendment defense introduce a federal issue. However, under the Mottley rule, the introduction of a federal issue in the defense does not confer federal question jurisdiction if the initial claim did not arise under federal law. \textcolor{blue}{(Correct)}\\
\\{[}4{]} Examining the Counterclaim: Garces files a counterclaim under the Federal Whistleblower Act, which is a federal statute. This counterclaim does indeed arise under federal law. \textcolor{blue}{(Correct)}\\
\\{[}5{]} Jurisdiction Based on Counterclaim: Federal courts can have supplemental jurisdiction over claims that are related to claims in the action which the court has original jurisdiction. Since Garces’s counterclaim arises under federal law, the court has original jurisdiction over the counterclaim and could potentially exercise supplemental jurisdiction over the related libel claim. \textcolor{blue}{(\textbf{Misinterpretation - Incompletely applying a legal rule}) - This step incorrectly applies supplemental jurisdiction principles. A counterclaim that arises under federal law does not retroactively establish original jurisdiction for the plaintiff's state law claim. Under Mottley, jurisdiction is assessed based on the plaintiff's claim, and the counterclaim cannot create subject matter jurisdiction for the law libel claim.} \\
\\{[}6{]} Assessing the Motion to Dismiss: Garces moves to dismiss for lack of subject matter jurisdiction. Given that his counterclaim arises under federal law, and considering the potential for supplemental jurisdiction over the libel claim, dismissing the case for lack of subject matter jurisdiction would be inappropriate. \textcolor{blue}{(\textbf{Misinterpretation - Misunderstanding the legal rules \& Misunderstanding the legal situation/issue at hand}) - This step misinterprets the legal rule and scenario. The federal counterclaim does not save the case from dismissal because the court’s jurisdiction is determined based on the plaintiff’s well-pleaded complaint.}\\ \\
{[}Final Answer with the correct option{]} \\
D. denied, because Garces’s counterclaim arises under federal law. \textcolor{blue}{(Wrong Conclusion from False Premise)}
\\\bottomrule
\end{tabular}
\caption{Example of another Misinterpretation and Wrong Conclusion from False Premise. This scenario involves interpreting federal jurisdiction rules, distinguishing between federal defenses and federal claims, and determining the impact of counterclaims on jurisdiction.}
\label{table: misinterpretation-2}
\end{table*}

\begin{table*}[h]
\footnotesize
\centering
\begin{tabular}{p{0.9\linewidth}}
\toprule
\textbf{Legal Context:}\\ The basic standard for removal jurisdiction is simple and sensible: a defendant sued in state court may remove a case to federal court if it could have been filed originally in federal court. The rationale for removal is to give both parties access to federal court for cases within the federal court’s jurisdiction. So it makes sense that the standard for removing a case to federal court should mirror the standard for filing one in federal court to begin with. So, 28 U.S.C. §1441(a) provides that ‘‘any civil action brought in a State court of which the district courts of the United States have original jurisdiction’’ may be removed to federal court. Consequently, the first question to ask in analyzing a removal problem is whether the claim asserted by the plaintiff fits into some category of federal jurisdiction. Is it a diversity case, or an arising-under case? A case between a citizen and foreign citizen, or some other case provided for in Article III, §2? If not, it can’t be removed. If it is, then, with a few minor exceptions, it can be. Here’s a first example to illustrate this relation between original and removal jurisdiction.          \\\hline
\textbf{Question:}\\ Castor Chemical Company, a California company, sues Pollux, from California, in state court for breach of contract, claiming that Pollux agreed to manufacture a drug for Castor and then reneged. Pollux answers the complaint, admitting that he refused to deliver the drug, and offering as an affirmative defense that he learned, after making the contract, that the drug was covered by a patent, so that, under federal patent law, he could not manufacture it without a license to do so. After answering, Pollux removed the action to federal court.\\ \\ \textbf{Options:}\\ A. The action is properly removed, because at the time he was removed, Pollux had raised a federal law issue in his answer. \\ B. The action is properly removed, because the case ‘‘arises under’’ federal law within the meaning of that phrase in Article III, §2. \\ C. The action is not properly removed, because it is a federal question case, not a diversity case. \\ D. The action is not properly removed, because the case does not ‘‘arise under’’ federal law within the meaning of 28 U.S.C. §1331.  \\\hline
\textbf{Human Annotator Analysis of Reasoning Chain 2:}\\ 1. The first step in determining whether a case can be removed to federal court is to identify whether the claim fits into any category of federal jurisdiction. In this case, the claim is for breach of contract between two California companies. \textcolor{blue}{(Factual Hallucination) - The breach of contract is between a person and a company.}\\ 2. The federal question jurisdiction, as per 28 U.S.C.§1331, applies to civil actions arising under federal law. In this case, Pollux has raised a federal law issue in his answer by asserting that he could not manufacture the drug due to a patent. \textcolor{blue}{(Correct)} \\ 3. However, the removal jurisdiction, as per 28 U.S.C. §1441(a), applies to civil actions brought in state court that have original jurisdiction in federal courts. \textcolor{blue}{(Correct)}\\ 4. In this case, the claim is not originally brought in federal court, but rather in state court. Therefore, even though a federal law issue has been raised, the case cannot be removed based on federal question jurisdiction alone.\\ \textcolor{blue}{(\textbf{Misinterpretation - Incorrectly applying the legal rule}) - The actual reason for why the case can not be removed to the federal court is that the plaintiff has invoked a state law tort in the claim and hence a federal court has no jurisdiction on this case even though the defendant raises a federal law issue. The correct standard is whether the plaintiff's original claim arises under federal law, not whether a federal issue is raised in the defense.}\\ 5. The correct answer is D: The action is not properly removed, because the case does not arise under federal law within the meaning of 28 U.S.C.§1331. \textcolor{blue}{(Correct Conclusion from False Premises)}\\ \\ \textcolor{blue}{(Both entities in same state jurisdiction -\textgreater Here, defendant is raising the federal jurisdiction claim-\textgreater Motley’s ‘arise under’ won’t apply (it only applies when plaintiff does that) -\textgreater Case can not be removed to Federal court)}\\ \bottomrule
\end{tabular}
\caption{Example of Factual Hallucination and Misinterpretation. The complexities of this scenario would include: applying the well-pleaded complaint rule to determine whether a case arises under federal law; distinguishing between state law and federal law claims for removal purposes; and recognizing that federal defenses do not confer federal jurisdiction, as established in the Mottley rule.}
\label{table: misinterpretation-3}
\end{table*}

\begin{table*}[h]
\footnotesize
\centering
\begin{tabular}{p{0.9\linewidth}}
\toprule
\textbf{Legal Context:}\\ Let’s start with the role of the United States Constitution in defining the limits of a court’s power to subject a defendant to jurisdiction. The Fourteenth Amendment bars a state from depriving a person of life, liberty or property ‘‘without due process of law,’’ that is, without a basically fair procedure. If it’s a court that’s doing the depriving—by entering a judgment against a person and forcing her to pay it—basic fairness requires that the defendant have some relationship to the state where the court sits that will make it fair to conduct the litigation there. In civil procedure terms, that means that the court must have ‘‘a basis to exercise personal jurisdiction’’ over the defendant. In the major personal jurisdiction cases, such as International Shoe, World-Wide Volkswagen, Daimler, Asahi, Burnham v. Superior Court, and Bristol-Myers Squibb, the United States Supreme Court has provided some guidance as to the types of relations to a state that will support the exercise of jurisdiction. Some relations that satisfy due process under the Fourteenth Amendment include domicile in a state,1 being ‘‘at home’’ in a state,2 minimum contacts that give rise to the claim,3 and service of process on an individual in the forum state. 4 That’s not an exhaustive list, and of course there are refinements and ambiguities, but the point is that the Supreme Court has upheld certain relations to a state as sufficient under the Fourteenth Amendment to support personal jurisdiction over a defendant there. However, it is not enough that a defendant has a contact with the state that is sufficient under the Due Process Clause to allow the state to exert jurisdiction over him... \textcolor{blue}{\textbf{...Truncated}} \\ \hline
\textbf{Question:}\\ Perrone, an Acadia resident, sues Margules, a Montanan, in an Acadia state court, for defamation. He claims that Margules called Acme Corporation, an Acadia company that was planning to hire Perrone, and made false and defamatory statements about his professional abilities that led Acme to decide not to hire him. Margules placed the call from her office in Montana. Assume that Acadia has a long-arm statute that authorizes personal jurisdiction over a defendant who ‘‘causes personal injury in this state by a tortious act or omission while physically present within this state.’’ Assume that this is the only statute that might authorize the Acadia court to exercise personal jurisdiction over Perrone for this claim.\\ \\ \textbf{Options:}\\ A. The Acadia statute authorizes the court to exercise jurisdiction over Margules in this case. However, it would be unconstitutional under the Fourteenth Amendment for the court to exercise jurisdiction over her on the facts of this case. \\ B. The Acadia statute does not authorize the court to exercise jurisdiction over Margules in this case. In addition, it would be unconstitutional under the Fourteenth Amendment for the court to exercise jurisdiction over her on the facts of this case. \\ C. The Acadia statute does not authorize the court to exercise jurisdiction over Margules in this case. However, it would be constitutional under the Fourteenth Amendment for the court to exercise jurisdiction over her on the facts of this case. \\ D. Although the Acadia statute does not authorize the court to exercise jurisdiction over Margules in this case, the court may do so because the claim arises out of her deliberate contacts with Acadia that gave rise to Perrone’s claim.                      \\ \hline
\textbf{Human Annotator Analysis of LLM generated Reasoning Chain:}\\ {[}1{]} The first step is to determine whether the Acadia statute authorizes the court to exercise jurisdiction over Margules in this case. The statute in question authorizes personal jurisdiction over a defendant who causes personal injury in Acadia by a tortious act or omission while physically present within the state. \textcolor{blue}{(Correct)}\\ \\ {[}2{]} In this case, Margules is not physically present in Acadia when she allegedly committed the tortious act. She made the defamatory statements from her office in Montana. \textcolor{blue}{(Correct)}\\ \\ {[}3{]} Therefore, based on the text of the statute, it does not appear to authorize the court to exercise jurisdiction over Margules in this case.\textcolor{blue}{(Correct)}\\ \\ {[}Final Answer: B. The Acadia statute does not authorize the court to exercise jurisdiction over Margules in this case.{]}\textcolor{blue}{(Wrong Conclusion from Incomplete Premises) - Making a decision based on an incomplete set of information, without considering all necessary legal aspects before jumping to the conclusion. Here the validity of the due process clause is not carefully examined and a premature conclusion is drawn.}\\ \\ However, it's important to note that even if the statute does not explicitly authorize jurisdiction in this case, a court may still exercise jurisdiction if the defendant has sufficient contacts with the forum state such that the exercise of jurisdiction comports with due process. This is known as the "minimum contacts" standard, which is a constitutional requirement separate from the statutory authorization. In this case, Margules' contact with Acadia through her defamatory statements may be sufficient to establish minimum contacts, but that analysis is beyond the scope of this question. \textcolor{blue}{(\textbf{Misinterpretation - Wrong assumptions derived from the provided context}) - No, that analysis is not outside the scope of this question!, in fact the expert has used “minimum” contact (instead of physical contact) to narrow the option to C. Margules' contact with Acadia through her defamatory statements is indeed be sufficient to establish minimum contacts and constitutionally}\\                 \\ \bottomrule
\end{tabular}
\caption{Example of Wrong Conclusion from Incomplete Premises. (The legal context has been truncated to save space.)}
\label{table: misinterpretation-4}
\end{table*}

\begin{table*}[h]
\footnotesize
\centering
\begin{tabular}{p{0.9\linewidth}}
\toprule
\textbf{Legal Context:}\\ 
The Due Process Clause of the Fourteenth Amendment prohibits a state from depriving a person of life, liberty, or property without due process of law. When a court enters a civil judgment against a person, it begins the process of taking the person’s property (usually in the form of money). Consequently, the due process clause requires the court to use a fair procedure in entering judgment. Certainly, one component of a fair procedure is to tell the defendant that the court is going to adjudicate her rights. Hence, the Due Process Clause requires a court to use a constitutionally adequate means of notifying the defendant that a lawsuit has been commenced against her. Typically, it is the plaintiff who does the legwork of serving process on the defendant. The statutes or court rules in every state contain detailed provisions governing how this notice of a lawsuit is provided to the defendant. Court rules may authorize various means of serving process. The most obvious is to deliver the initiating papers in the case to the defendant in person, called ‘‘personal service of process.’’ Other methods are often authorized as well, though they are less certain to actually inform the defendant about the suit. The service rules may provide, for example, that the papers may be left with someone at the defendant’s home or place of business or slipped under the door. Some authorize service to the defendant by certified or ordinary mail. In some circumstances, court rules or statutes may authorize service by publication, by printing a notice of the suit in the legal notices in the local newspaper for several weeks. Such publication notice is sometimes referred to as ‘‘constructive service,’’ as opposed to actual delivery of the papers to the defendant. Which of these methods is constitutionally adequate under the Due Process Clause? Justice Jackson’s wise opinion in Mullane v. Central Hanover Bank \& Trust Co., 339 U.S. 306 (1950) does not provide a mechanical answer to that question, but instead establishes broad standards for constitutionally adequate notice. Mullane involved a device called a common trust fund, which allowed a bank to pool a group of small trusts into one common fund, which could be managed more efficiently. Income (or losses) from the trust would then be shared by each smaller trust in proportion to their contributions to the fund. Periodically, the trustee of the fund would file accounts with the court. If the court approved the accounts, its judgment would bar investors from suing the bank for mismanagement. Since this judicial accounting could deprive trust beneficiaries of property—the right to sue the bank for mismanagement—due process required adequate notice to them of the proceeding. The problem in Mullane was that it was difficult to ascertain who all the beneficiaries were. Beneficiaries currently entitled to receive income from the trust were known, and the bank had their addresses in its files. But most trusts have contingent beneficiaries, such as the children or heirs of a beneficiary if the primary beneficiary dies. The bank might not know who these beneficiaries were, and they could change periodically. Perhaps, with a large investment of time by lawyers and investigators, the bank could identify most them, but doing so would eat up the profits of the common trust fund.\textcolor{blue}{\textbf{...Truncated}} \\ \midrule
\textbf{Question:}\\ In Mullane v. Central Hanover Bank \& Trust Co., the Supreme Court held that                              \\ \midrule
\textbf{Human Annotator Analysis of LLM generated Reasoning Chain:}\\ {[}1{]} The case of Mullane v. Central Hanover Bank \& Trust Co. involved a common trust fund where the bank pooled multiple trusts into one fund for more efficient management. The trustee would file accounts with the court, and if approved, the judgment would bar investors from suing the bank for mismanagement, potentially depriving them of their property right to sue. The issue was who the bank had to notify about the suit and how. \textcolor{blue}{(Correct)}\\

{[}2{]} The Supreme Court held that due process requires notice reasonably calculated to apprise interested parties of the pendency of the action and afford them an opportunity to present their objections. What is reasonable in one circumstance may not be in another. In Mullane, several factors reduced the risk of error if notice failed to reach all beneficiaries: the appointment of a guardian to represent their interests, and the likelihood that many beneficiaries would receive notice and share the interests of all beneficiaries in the settlement of the accounts. \textcolor{blue}{(Correct)}\\

{[}3{]} Based on the information provided in the case, the Court did not require that each individual with an interest in the fund be given personal notice of the action. Instead, the Court focused on the reasonableness of the notice given under the circumstances.\textcolor{blue}{(Correct)}\\

{[}4{]} Therefore, the correct answer is D: due to the large number of persons whose interests might be affected by the proceedings, notice by publication was sufficient on the facts of the case. \textcolor{blue}{(Wrong Conclusion from Incomplete Premises) - the LLM assumed on its own that notice by publication is sufficient, whereas the court ruling in Mullane, even though focusing on the reasonableness of the notice given in such circumstances, never mentioned that notice by publication was the right way! The ruling went for representation by Guardianship and Common Interests representation)
Step [4] simplifies the Court's nuanced position into a broad and incorrect generalization. While the Court did acknowledge the use of publication notice as part of the process, it did not conclude that publication alone was sufficient under the circumstances of this specific case.}\\\\

[Final Answer: D] 
                                                            \\ \bottomrule
\end{tabular}
\caption{Example of Wrong Conclusion from Incomplete Premises.}
\end{table*}

\begin{table*}[h]
\footnotesize
\centering
\begin{tabular}{p{{0.9\linewidth}}}
\toprule
\textbf{Legal Context:}\\ The relation between statutory diversity jurisdiction and the constitutional grant Article III, §2 of the Constitution creates diversity jurisdiction in very general terms. It does not expressly state whether there must be ‘‘complete diversity,’’ or whether diversity might be satisfied if some plaintiffs are diverse from some defendants.  Strawbridge v. Curtiss,  of course, held that all defendants must be diverse from all plaintiffs, but for many years it was unclear whether  Strawbridge interpreted Article III or interpreted the statute by which Congress conveys diversity jurisdiction to the lower federal courts. If  Strawbridge interpreted the language ‘‘citizens of different states’’ in Article III, the complete diversity rule would be constitutionally required, and could not be changed  \textcolor{blue}{\textbf{...Truncated}}                  \\ \hline
\textbf{Question:}\\ A diversity case cannot be heard in federal court unless the amount in controversy is at least\\ \hline 
\textbf{Options:}\\ A.  \$75,000, counting interest and costs as well as the plaintiff’s damages.\\ B.  \$75,000, not counting interest and costs.\\ C.  \$75,000.01, counting interest and costs. \\ D.  \$75,000.01, exclusive of interest and costs.
\\ \hline
\textbf{Human Annotator Analysis of LLM generated Reasoning Chain:}\\ {[}1{]} The legal context explains that the amount-in-controversy requirement is a condition imposed by Congress on the exercise of diversity jurisdiction in federal courts, which is not present in the constitutional grant of diversity jurisdiction in Article III, §2. \textcolor{blue}{(Correct)}\\ \\ {[}2{]} The question asks for the minimum amount in controversy that must be present for a diversity case to be heard in federal court. \textcolor{blue}{(Correct)}\\ \\ {[}3{]} Based on the legal context, the constitutional grant of diversity jurisdiction in Article III, §2, does not contain any monetary restriction. Therefore, there is no minimum amount in controversy required for a case to be considered a diversity case under the Constitution. \textcolor{blue}{(Irrelevant Premise) - The question is about the statutory requirement, not the constitutional one.}\\ \\ {[}4{]} However, the federal courts derive their jurisdiction from Congress, and Congress has imposed an amount-in-controversy requirement as a condition for the exercise of diversity jurisdiction. \textcolor{blue}{(Correct)}\\ \\ {[}5{]} Therefore, the correct answer is [B]: \$75,000, not counting interest and costs. This is the minimum amount in controversy required for a diversity case to be heard in federal court under the statutory grant of jurisdiction \textcolor{blue}{(Wrong Conclusion from False Premises) -  Mistral doesn’t fully understand the terms of the legal context, which states “the amount in controversy exceeds the sum or value of \$75,000, exclusive of interest or costs” not equals or exceeds, meaning the value must be greater than, not equal to, \$75,000.}\\ \bottomrule
\end{tabular}
\caption{Example of an Irrelevant Premise.}
\label{table:human_annotation_example_final}
\end{table*}
\end{document}